\documentclass{article} %
\usepackage{iclr2027_conference,times}

\usepackage{amsmath,amsfonts,bm}

\def\eqref#1{equation~\ref{#1}}

\def\1{\bm{1}}

\DeclareMathAlphabet{\mathsfit}{\encodingdefault}{\sfdefault}{m}{sl}
\SetMathAlphabet{\mathsfit}{bold}{\encodingdefault}{\sfdefault}{bx}{n}

\usepackage{hyperref}
\usepackage{url}
\usepackage{booktabs}
\usepackage{tabularx}
\usepackage{array}
\usepackage{graphicx}
\usepackage{placeins} %
\usepackage{pdflscape} %

\usepackage{tikz}
\usetikzlibrary{arrows.meta,calc,fit}
\definecolor{figPol}{HTML}{534AB7}   %
\definecolor{figPolBg}{HTML}{EEEDFE}
\definecolor{figPolTx}{HTML}{3C3489}
\definecolor{figRef}{HTML}{0F766E}   %
\definecolor{figRefBg}{HTML}{E1F5EE}
\definecolor{figRefTx}{HTML}{085041}

\usepackage{booktabs}
\usepackage{array}
\usepackage{tabularx}
\usepackage[table]{xcolor}
\newcolumntype{Y}{>{\raggedright\arraybackslash\setlength{\parindent}{0pt}}X}
\newcolumntype{Z}{>{\raggedright\arraybackslash}X}

\definecolor{cInstr}{HTML}{1F2937}   %
\definecolor{cCtx}{HTML}{0F766E}     %
\definecolor{cOut}{HTML}{9A3412}     %
\newcommand{\pInstr}[1]{\textcolor{cInstr}{#1}}
\newcommand{\pCtx}[1]{\textcolor{cCtx}{\textbf{#1}}}
\newcommand{\pOut}[1]{\textcolor{cOut}{\texttt{#1}}}
\newcommand{\slot}[1]{\textcolor{cCtx}{\textbf{$\langle$\textit{#1}$\rangle$}}}
\newcommand{\legendbox}[1]{\textcolor{#1}{\rule{0.7em}{0.7em}}}

\usepackage{multirow}
\usepackage{comment}

\iclrfinalcopy

\title{GrammarRL: Effective Grammar-Constrained Decoding via Reinforcement Learning}

\author{
Gabriele Tuccio$^{1,3}$ \quad
Antonino Furnari$^{1}$ \quad
Aldo Gangemi$^{2,3}$ \quad
Misael Mongiovì$^{1,3}$
\\
$^{1}$ University of Catania \qquad
$^{2}$ University of Bologna \qquad
$^{3}$ 
ISTC - National Research Council, Italy
}

\begin{document}

\maketitle
\begin{abstract}
Grammar-constrained generation guarantees syntactic validity, but can substantially degrade semantic quality when the model’s preferred outputs are poorly aligned with the imposed grammar. This trade-off is particularly severe when the prompt is underspecified or the model has limited instruction-following ability. Beam search can partially mitigate these failures by exploring multiple valid sequences, but its computational cost grows with beam width, while sequence-level probability is only an imperfect proxy for semantic quality.

We introduce GrammarRL, a label-free reinforcement learning method that adapts language models to grammar constraints without requiring annotated data. GrammarRL optimizes the model using two complementary self-supervised rewards derived from its own likelihoods: a direct reward, measuring how likely the constrained output is given the input, and a reverse reward, measuring how well the input can be reconstructed from the generated output. We optimize these rewards with a Reinforce Leave-One-Out (RLOO) objective over groups of grammar-constrained rollouts, augmented with the top-1 beam-search hypothesis and regularized towards a frozen base model.

We evaluate GrammarRL on sign language gloss translation, hierarchical text classification, and named entity recognition using Llama models ranging from 1B to 8B parameters. GrammarRL consistently outperforms constrained greedy decoding, with an average improvement of 9.8 points and gains of up to 22.8 BLEU. It matches or outperforms beam search on two of the three tasks while preserving greedy-decoding inference cost. Ablations further show that the two rewards are complementary: either reward alone can underperform the untrained baseline, whereas their combination consistently improves upon it.
\end{abstract}

\section{Introduction}

\begin{figure}[t]
    \centering
    \includegraphics[width=\linewidth]{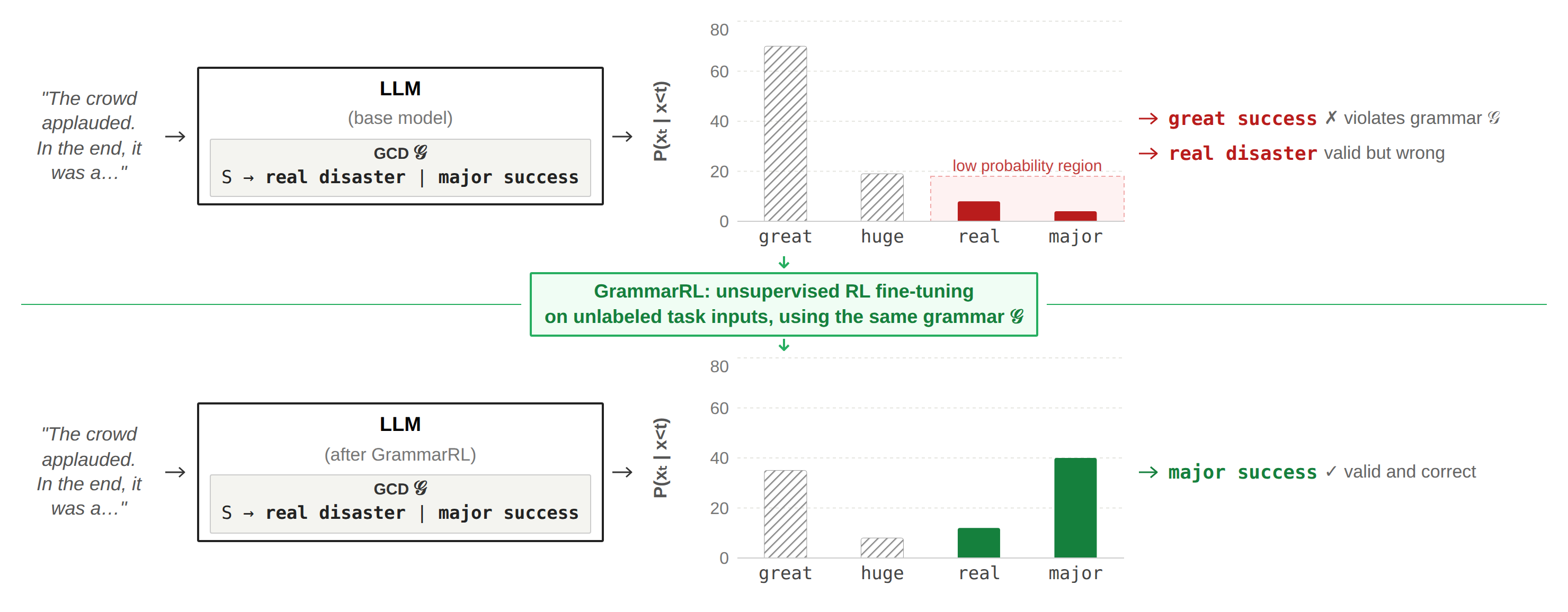}
    \caption{ Illustration of grammar-constrained decoding pitfalls and the GrammarRL solution. \textbf{Top:} The base model assigns most of its probability mass to invalid tokens (e.g., ``great'', ``huge''). Constrained to a low-probability region where differences are narrowed, a greedy step chooses ``real'', leading to the valid but incorrect sequence ``real disaster''. \textbf{Bottom:} Unsupervised RL fine-tuning with GrammarRL reshapes the distribution on unlabeled task inputs. The model directly assigns high probability to valid and correct continuations (``major success''), bypassing the need for complex inference-time search.}
    \label{fig:grammar_rl}
\end{figure}

Grammar-constrained decoding is widely used to guarantee the formal correctness of generated outputs \citep{geng-etal-2023-grammar,dong2025xgrammar,tuccio-etal-2025-grammar,beurerkellner2024domino,ugare2025syncode,willard2023efficient}.
By masking invalid tokens at every generation step, a grammar restricts the model to a valid subset of the output space and prevents syntactically malformed sequences from being produced. However, formal validity alone does not guarantee that the generated sequence expresses the desired semantics \citep{tam-etal-2024-speak,banerjee2025crane}.
A model may assign high probability to a valid token that leads to an undesirable continuation, creating a ``trap'' in which subsequent greedy decisions remain syntactically valid but semantically wrong  \citep{park2024grammar,biagiola2026alignmentproblemconstrainedcode}. As illustrated in the top half of Figure \ref{fig:grammar_rl}, a base model might assign most of its probability mass to tokens outside the grammar. When forced to select from the remaining valid options in a low-probability region, a greedy choice can commit the model to a syntactically valid but contextually incorrect continuation.

This problem is particularly important in \emph{data-scarce} settings, where annotated target sequences are limited or unavailable to guide model generation. The standard remedy is inference-time search. Beam search maintains several candidate prefixes and can recover sequences that greedy decoding misses \citep{lu-etal-2022-neurologic}, but it is primarily a \emph{mode-seeking} strategy, and high sequence-level probability does not necessarily translate into high-quality outputs \citep{holtzman2019curious}. Posterior-approximation methods instead target the grammar-conditioned distribution over valid sequences, for instance by exploring multiple valid continuations and reweighting them with importance sampling \citep{lew2023smc,park2024grammar,loula2025syntactic}. Both families, however, obtain their improvements through additional computation at every inference call.

We instead ask whether useful sequence-level preferences can be learned \emph{before inference} in the data-scarce, label-free setting, using only signals available without annotated target outputs. We introduce \emph{GrammarRL}, a reinforcement learning framework that uses grammar-constrained rollouts and two complementary rewards derived from a frozen pretrained model to shift probability mass toward continuations that are both grammatical and correct (see the bottom half of Figure \ref{fig:grammar_rl}). The \emph{direct} reward measures the conditional log-likelihood %
of a candidate output under the frozen reference model,
and encourages the policy to discover valid sequences that are likely under the reference model. We also define a \emph{reverse} reward, since high output likelihood alone does not necessarily identify task-useful outputs. In particular, among valid candidates with similar likelihood, some may progressively discard information from the input and converge toward generic or degenerate sequences. We therefore complement the direct reward with a \emph{reverse} reward, %
which favors outputs that retain sufficient information about the source for the reference model to recover it. The two rewards thus capture complementary properties of a desirable output: output plausibility under the frozen reference model and preservation of source-relevant information. Both signals are available without reference outputs, providing sequence-level supervision in the label-free setting.
Because these rewards are computed from discrete sequences generated under the grammar constraint, they cannot be optimized by backpropagating through the decoding decisions. GrammarRL therefore treats constrained generation as a policy and optimizes the expected reward with reinforcement learning. Specifically, we use a Reinforce Leave-One-Out (RLOO) objective over groups of grammar-constrained rollouts, using the rewards of the other samples in each group to form a relative baseline for each candidate. We additionally include the top-1 beam-search hypothesis as a candidate during training, allowing beam search to provide useful trajectories without turning the objective into beam-search imitation. The policy is trained to increase the probability of high-reward sequences while remaining regularized toward the frozen reference model.

This reinforcement learning formulation is central to GrammarRL. Rather than improving posterior approximation at inference time or explicitly maximizing the sequence-level probability, we use model-derived sequence-level rewards to transfer sequence-level preferences into the model parameters, amortizing part of the inference-time correction during a label-free tuning phase. After training, these preferences can be exploited with greedy constrained decoding, without maintaining multiple hypotheses or performing iterative correction.

We evaluate GrammarRL on three structured generation tasks---Sign Language Translation (Gloss), Hierarchical Classification (WoS), and Named Entity Recognition (CoNLL)---and compare it with beam search and with sampling-based posterior approximation. The results suggest that sequence-level preferences can be learned from unlabeled inputs and internal model signals, shaping the model toward informative and task-relevant constrained outputs without costly search at inference time.
Our contributions are as follows:
\begin{itemize}
    \item We introduce GrammarRL, a reinforcement learning method that adapts language models to grammar
    constraints without annotated target outputs, significantly improving the output quality without increasing inference cost. %
    \item We propose a bidirectional reward computed by a frozen reference model, which combines the
    likelihood of the output given the input with the likelihood of reconstructing the input from the output.
    We show that the two terms are complementary: each one alone favors a different kind of degenerate output,
    whereas %
    their combination improves over greedy decoding of the
    frozen baseline in every setting.
    \item We show on three structured-generation tasks and three model sizes that GrammarRL improves
    constrained greedy decoding by 9.8 points on average and matches or exceeds beam search on two of the
    three tasks, at greedy-decoding inference cost. On Gloss and CoNLL it also matches or outperforms sampling-based
    posterior approximation (importance sampling and SMC) that uses ten times more generation.
\end{itemize}

\section{Related Work}
\label{sec:related_work}

\textbf{Grammar-constrained decoding (GCD)} is widely used to guarantee that language-model outputs satisfy formal requirements, including programming languages, SQL, regular expressions, and structured formats such as JSON. Early approaches enforce constraints incrementally by rejecting invalid continuations \citep{scholak2021picard,poesia2022synchromesh}, while subsequent methods improve the generality and efficiency of grammar-guided generation. Outlines supports regular expressions and context-free grammars \citep{willard2023efficient}, SynCode provides sound and complete CFG-based masking \citep{ugare2025syncode}, DOMINO addresses the mismatch between grammar terminals and subword vocabularies \citep{beurerkellner2024domino}, and XGrammar combines precomputation with systems-level optimizations for efficient CFG-constrained serving \citep{dong2025xgrammar}. 
Grammar-LLM introduces an \emph{LL(prefix)} grammar notation, equivalent to LL(1) grammars, for efficient constrained decoding in LLMs, enabling expressive constraints with linear-time syntactic enforcement \citep{tuccio-etal-2025-grammar}.
While these methods make constrained decoding increasingly efficient and general, they primarily treat constraints as an inference-time mechanism and do not explicitly account for how local masking decisions affect the model's global distribution or task-relevant information.

\textbf{Posterior approximation for constrained generation} addresses this limitation by viewing constrained decoding as inference under a validity condition. Rather than relying solely on local token admissibility, recent methods incorporate information about future valid completions. Grammar-Aligned Decoding (GAD) estimates future grammaticality to guide generation \citep{park2024grammar}, while Sequential Monte Carlo methods formulate constrained generation as probabilistic inference and approximate the target distribution with multiple particles \citep{lew2023smc,loula2025syntactic}. Other approaches construct globally constrained proposal distributions \citep{dang2026mitigating}, use importance weighting and resampling to correct local sampling bias \citep{ahmed2025controllable}, explicitly estimate future validity \citep{nie2026future}, or revise prefixes through backtracking when later conflicts arise \citep{li2026adaptrack}. These methods reduce the local myopia of hard masking, but they do so by introducing additional computation at inference time while leaving the underlying model unchanged. Our work instead asks whether the same sequence-level signals can be transferred into the model during training, thereby reducing the need for inference-time search or distribution correction.

\textbf{Learning from formal constraints} explores a complementary direction: using constraints themselves as supervision when conventional labels are unavailable. Prior work combines grammar information with reinforcement learning for neural program synthesis \citep{bunel2018leveraging}, while unsupervised semantic parsing uses grammar-constrained generation together with paraphrasing objectives to learn without paired logical forms \citep{wu2021paraphrasing}. More recently, Schema Reinforcement Learning (SRL) uses a fine-grained JSON-schema validator as a reinforcement signal to improve schema-compliant generation \citep{lu2025schema}. These approaches show that formal constraints can provide useful training signals without reference outputs, but their objectives primarily reward structural or schema compliance. 
GrammarRL instead aims at semantic quality under grammar-constrained decoding. Rather than contrasting valid and invalid sequences, which gives a weak signal when most samples are invalid, it optimizes the model directly within the grammar-constrained space, where every sequence is valid by construction, so the training signal focuses on semantic quality.

\section{Methodology}
\label{sec:methodology}

\subsection{Preliminaries}
\label{subsec:preliminaries}

\paragraph{Grammar-constrained decoding.}
We consider autoregressive generation under a formal constraint, with formal grammars as the primary case. Let $G$ denote a grammar and let $\mathcal{L}(G)$ be the set of valid output sequences. More generally, the same formulation applies to any constraint for which the set of admissible next tokens can be determined incrementally from the generated prefix. Let $\mathcal{M}_{G}(y_{<t}) \subseteq \mathcal{V}$ denote the set of tokens that can be appended to a prefix $y_{<t}$ while remaining extendable to a sequence in $\mathcal{L}(G)$. \footnote{For a context-free grammar, this set is computable in polynomial time; for deterministic context-free grammars, such as LL or LR grammars, $\mathcal{M}_{G}(y_{<t})$ can be maintained incrementally in time linear in the length of the generated sequence.}

Given an unconstrained language-model policy $\pi_{\theta}$, grammar-constrained decoding masks tokens outside $\mathcal{M}_{G}(y_{<t})$ and renormalizes the remaining probability mass:

\begin{equation}
\label{eq:constrained-policy}
\pi_{\theta}^{G}(y_t \mid x, y_{<t})
=
\frac{
\pi_{\theta}(y_t \mid x, y_{<t})
\mathbb{I}\!\left[
y_t \in \mathcal{M}_{G}(y_{<t})
\right]
}{
\displaystyle
\sum_{v \in \mathcal{M}_{G}(y_{<t})}
\pi_{\theta}(v \mid x, y_{<t})
}.
\end{equation}

This procedure guarantees formal validity, but the decision at each step remains local. In particular, token admissibility only indicates whether a prefix can still be completed into a valid sequence; it does not directly account for the probability that the model assigns to the valid continuations that follow each admissible token. The ideal model distribution conditioned on the constraint is instead
\begin{equation}
\label{eq:grammar-posterior}
p_{\theta}(y \mid x, G)
=
\frac{
p_{\theta}(y \mid x)\,
\mathbb{I}\!\left[y \in \mathcal{L}(G)\right]
}{
\displaystyle
\sum_{y' \in \mathcal{L}(G)}
p_{\theta}(y' \mid x)
}.
\end{equation}
Computing this distribution exactly requires accounting for the total probability of complete valid continuations and is generally impractical for autoregressive generation. This gap motivates inference-time strategies such as beam search \citep{lu-etal-2022-neurologic} and posterior-approximation methods \citep{lew2023smc,loula2025syntactic} that explore multiple candidates or estimate future validity.

In GrammarRL, this constrained generation process defines the policy used during reinforcement learning. We denote the trainable policy by $\pi_{\theta}$ and the frozen pretrained model by $\pi_{\mathrm{ref}}$.
The policy generates grammar-constrained candidates and is updated from their sequence-level rewards, while $\pi_{\mathrm{ref}}$ provides a fixed reference for KL regularization during optimization.

\subsection{GrammarRL}
\label{subsec:grammarl}

\begin{figure}[t]
\centering
\begin{tikzpicture}[
    font=\small,
    >={Stealth[length=5pt,width=4pt]},
    arr/.style={->, draw=black!65, line width=0.6pt},
    box/.style={draw=black!45, fill=black!4, rounded corners=3pt, line width=0.4pt,
                align=center, inner sep=3pt},
    pol/.style={box, draw=figPol, fill=figPolBg, text=figPolTx},
    ref/.style={box, draw=figRef, fill=figRefBg, text=figRefTx},
    cand/.style={box, minimum width=0.72cm, minimum height=0.55cm, inner sep=1pt},
    zone/.style={draw=black!35, dashed, rounded corners=4pt, line width=0.5pt},
    zlabel/.style={font=\small\bfseries, anchor=north west, text=black!80},
    note/.style={font=\scriptsize, text=black!60, align=center},
]
\draw[zone] (0,0) rectangle (4.25,4.4);
\draw[zone] (4.65,0) rectangle (9.15,4.4);
\draw[zone] (9.55,0) rectangle (13.9,4.4);
\node[zlabel] at (0.05,4.4) {Rollout};
\node[zlabel] at (4.70,4.4) {Scoring};
\node[zlabel] at (9.60,4.4) {Update};

\node[box, minimum width=2.2cm, minimum height=0.55cm] (x) at (2.12,3.55) {input $x$};
\node[pol, minimum width=3.3cm, minimum height=0.85cm] (pol) at (2.12,2.45)
     {policy $\pi_\theta^{G}$\\[-1pt]{\scriptsize LoRA + grammar mask}};
\node[cand] (c1) at (0.62,1.25) {$y^{(1)}$};
\node       (cd) at (1.40,1.25) {$\cdots$};
\node[cand] (cN) at (2.18,1.25) {$y^{(N)}$};
\node[cand, minimum width=0.95cm] (cB) at (3.45,1.25) {$y^{(N+1)}$};
\node[note] at (1.40,0.55) {$N$ sampled};
\node[note] at (3.45,0.55) {beam top-1};
\node[draw=black!35, rounded corners=3pt, line width=0.4pt, fit=(c1)(cB), inner sep=3pt] (cands) {};
\draw[arr] (x) -- (pol);
\draw[arr] (pol.south) -- (pol.south |- cands.north);

\node[ref, minimum width=3.9cm, minimum height=0.9cm] (dir) at (6.9,3.2)
     {direct reward\\[-1pt]$\log \pi_{\mathrm{ref}}(y \mid x)$};
\node[ref, minimum width=3.9cm, minimum height=0.9cm] (rev) at (6.9,1.9)
     {reverse reward\\[-1pt]$\log \pi_{\mathrm{ref}}(x \mid \tau(y))$};
\node[note] at (6.9,0.65) {frozen $\pi_{\mathrm{ref}}$, no grammar mask};
\draw[arr] (cands.east) -- (4.45,1.25) |- (dir.west);
\draw[arr] (4.45,1.9) -- (rev.west);

\node[box, minimum width=3.7cm, minimum height=0.9cm] (rw) at (11.72,3.2)
     {reward $r$\\[-1pt]{\scriptsize $\lambda$-weighted, $\sigma$-scaled}};
\node[box, minimum width=3.7cm, minimum height=0.7cm] (adv) at (11.72,2.0)
     {RLOO advantage $A^{(i)}$};
\node[pol, minimum width=3.7cm, minimum height=0.9cm] (upd) at (11.72,0.85)
     {update $\pi_\theta$\\[-1pt]{\scriptsize LoRA, KL to $\pi_{\mathrm{ref}}$}};
\draw[arr] ([yshift=5pt]dir.east) -- ([yshift=5pt]dir.east -| rw.west);
\draw[arr] (rev.east) -- (9.35,1.9) |- ([yshift=-7pt]rw.west);
\draw[arr] (rw) -- (adv);
\draw[arr] (adv) -- (upd);

\draw[arr, dashed, draw=black!50] (upd.south) -- (11.72,-0.45) -| (2.12,0);
\node[note, anchor=north] at (6.9,-0.5) {repeat with the updated policy};

\node[pol, minimum width=13.9cm, minimum height=0.55cm] at (6.95,-1.3)
     {test time: greedy decoding from $\pi_\theta^{G}$, no further exploration is needed};

\node[draw=figPol, fill=figPolBg, minimum size=7pt, inner sep=0pt] at (3.1,-2.0) {};
\node[note, anchor=west, text=black!75] at (3.25,-2.0) {trainable policy, grammar mask active};
\node[draw=figRef, fill=figRefBg, minimum size=7pt, inner sep=0pt] at (8.1,-2.0) {};
\node[note, anchor=west, text=black!75] at (8.25,-2.0) {frozen reference model, no grammar mask};
\end{tikzpicture}
\caption{Overview of GrammarRL framework. For each input $x$, the policy $\pi_\theta^{G}$ (LoRA adapters on
the frozen model, decoding with the grammar mask) generates $N$ sampled candidates and the top-1 beam-search
candidate. The frozen reference model $\pi_{\mathrm{ref}}$ scores every candidate without the grammar mask in
two directions: the direct reward measures how probable the candidate is given the input, and the reverse reward
measures how well the input can be reconstructed from the candidate. The combined reward defines a
leave-one-out advantage that updates the LoRA adapters, with KL regularization toward $\pi_{\mathrm{ref}}$. At
test time, GrammarRL uses only greedy grammar-constrained decoding.}
\label{fig:method}
\end{figure}
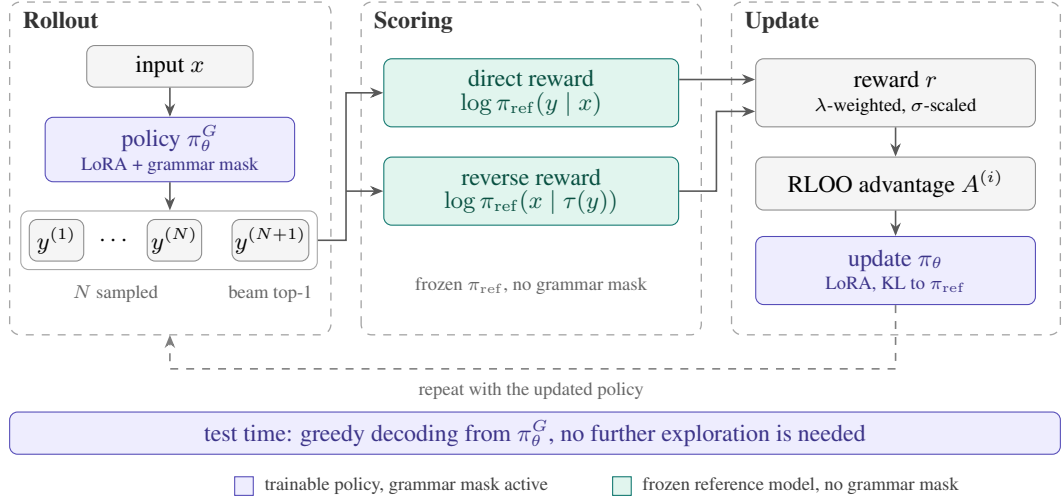

GrammarRL is a label-free reinforcement learning procedure.
A trainable policy generates grammar-valid candidates, a frozen reference model scores them with a bidirectional sequence-level reward, and the policy is updated with a leave-one-out policy gradient. Figure~\ref{fig:method} summarizes the procedure.

\vspace{-10pt}
\paragraph{Setup.}
GrammarRL involves two models. The \emph{reference model} $\pi_{\mathrm{ref}}$ is the frozen pretrained model:
it is never updated, and it is used to score candidates and to regularize training. The \emph{policy} $\pi_\theta$
is the model being trained: it shares the frozen weights of $\pi_{\mathrm{ref}}$ and adds trainable LoRA adapters
(Appendix~\ref{app:implementation-details}), so that $\pi_\theta = \pi_{\mathrm{ref}}$ at the start of training.

The grammar is used only to \emph{generate} candidates, which are sampled from the constrained policy
$\pi_\theta^G$ in \eqref{eq:constrained-policy} and are therefore always valid. Every sequence probability used
to score or train on these candidates is instead computed with the \emph{unconstrained} model,
\begin{equation}
\label{eq:unconstrained-seq}
\pi(y \mid x) = \prod_{t=1}^{|y|} \pi(y_t \mid x, y_{<t}),
\qquad \pi \in \{\pi_\theta, \pi_{\mathrm{ref}}\},
\end{equation}
without the renormalization over $\mathcal{M}_G$ in \eqref{eq:constrained-policy}. This holds for both rewards
and for both terms of the loss. The distinction matters because renormalization rescales the valid tokens to sum
to one at every step, so a valid sequence reached through tokens that the model considers unlikely can still
receive a high constrained probability. Unconstrained probabilities preserve this information, and candidates
that are valid but unlikely under the model are scored accordingly.
\vspace{-10pt}
\paragraph{Rollouts.}
For each input $x$, we generate a group of $N+1$ candidates: $y^{(1)}, \dots, y^{(N)} \sim \pi_\theta^G(\cdot \mid x)$, and $y^{(N+1)}$, the top-1 hypothesis of beam search (width $B$) under $\pi_\theta^G$.
Sampling explores the current policy, while beam search adds a candidate from a higher-probability region of the constrained output space.
The beam hypothesis is not a target: it is scored with the same reward as the sampled candidates and affects the update only through its relative reward.
\vspace{-10pt}
\paragraph{Bidirectional reward.}
Each candidate $y$ is scored by the frozen model $\pi_{\mathrm{ref}}$ in two directions.
The \emph{direct} term is the length-normalized log-likelihood of the candidate:
\begin{equation}
\label{eq:direct}
\mathcal{R}_{\mathrm{direct}}(x, y)
=
\frac{1}{|y|} \sum_{t=1}^{|y|} \log \pi_{\mathrm{ref}}(y_t \mid x, y_{<t}).
\end{equation}
It penalizes candidates that are valid under $G$ but unlikely under the pretrained model.

In the label-free setting, maximizing log-likelihood alone tends to favor generic outputs that discard information from $x$, or degenerate sequences (see Figure~\ref{fig:loglik-not-proxy} in Appendix~\ref{app:Qualitative} for an example).
The \emph{reverse} term addresses this by scoring the source given the candidate.
Let $\tau(\cdot)$ be a prompt template that swaps the roles of source and target:
\footnote{concrete examples of the forward and reverse prompt templates used for Gloss, WoS, and CoNLL are provided in Table~\ref{tab:prompt-system} in Appendix~\ref{app:Qualitative}}

\begin{equation}
\label{eq:reverse}
\mathcal{R}_{\mathrm{reverse}}(x, y)
=
\frac{1}{|x|} \sum_{k=1}^{|x|} \log \pi_{\mathrm{ref}}\!\left(x_k \mid \tau(y), x_{<k}\right).
\end{equation}
It is computed by teacher forcing in a single forward pass and is higher when $y$ preserves more information about $x$.

The reward is
\begin{equation}
\label{eq:reward}
r(x, y)
=
(1-\lambda)\,\frac{\mathcal{R}_{\mathrm{direct}}(x, y)}{\sigma_{\mathrm{direct}}}
+
\lambda\,\frac{\mathcal{R}_{\mathrm{reverse}}(x, y)}{\sigma_{\mathrm{reverse}}},
\qquad \lambda \in [0, 1],
\end{equation}
where $\sigma_{\mathrm{direct}}$ and $\sigma_{\mathrm{reverse}}$ are fixed constants, measured before training as the mean within-group standard deviation of each term under the frozen reference model on a held-out set of 100 unlabeled inputs per task (Appendix~\ref{app:calibration}, Table~\ref{tab:calibration-all}).

\vspace{-10pt}
\paragraph{Policy optimization.}
We use the Reinforce leave-one-out estimator (RLOO) \citep{Kool2019Buy4R}, which compares each candidate with the other $N$ in its group:
\begin{equation}
\label{eq:rloo-advantage}
A^{(i)}
=
r^{(i)} - \frac{1}{N} \sum_{j \neq i} r^{(j)},
\qquad
r^{(i)} = r(x, y^{(i)}).
\end{equation}
The advantage is positive when a candidate outperforms the other candidates generated for the same input and negative otherwise, which provides a learning signal without an annotated target. We use RLOO rather than Group Relative Policy Optimization (GRPO) \citep{Shao2024DeepSeekMath} because our rollout groups are small and combine sampled and beam-search trajectories: the leave-one-out baseline excludes the candidate's own reward while preserving the relative comparison needed for our label-free reward.
Since the reward is computed by the frozen model, the advantages do not depend on $\theta$.
The policy is updated by minimizing
\begin{equation}
\label{eq:loss}
\mathcal{L}(\theta)
=
\frac{1}{N+1} \sum_{i=1}^{N+1}
\Big[
- A^{(i)} \log \pi_\theta(y^{(i)} \mid x)
+ \beta\, \Delta^{(i)}
\Big],
\qquad
\Delta^{(i)}
=
\log \pi_\theta(y^{(i)} \mid x) - \log \pi_{\mathrm{ref}}(y^{(i)} \mid x),
\end{equation}
where $\beta \ge 0$.
The second term penalizes candidates that deviate substantially from the reference model.

\section{Experiments}
\subsection{Experimental Setup}
\label{sec:exp-setup}

We evaluate GrammarRL on three structured-generation tasks that differ
substantially in output space, structural constraints, and information
requirements: Sign language translation on ASLG-PC12 ~\citep{othman2012english},
hierarchical Web-of-Science (WoS) classification ~\citep{kowsari2017hdltex},
and CoNLL-2003 named-entity recognition ~\citep{tjong-kim-sang-de-meulder-2003}. All experiments use Grammar-LLM \citep{tuccio-etal-2025-grammar} as the grammar-constrained generation framework throughout training and inference. Prompts for all tasks are reported in Appendix~\ref{app:prompts}.

\textbf{Sign Language Translation.}
We consider English-to-gloss generation, with a vocabulary of 16{,}120 gloss
terminals and a grammar covering the output taxonomy. We report BLEU as the
evaluation metric~\citep{papineni2002bleu}.
The test set contains 2{,}000 examples. Each prompt includes 20
fixed few-shot examples, together with an input-specific subset of the gloss
taxonomy precomputed before training.

\textbf{Hierarchical Classification.}
For WoS, the model generates a JSON object containing the predicted domain and
area labels. The grammar constrains both fields to their closed vocabularies
and enforces the required structure. We report Hierarchical F1 on the
2{,}000-example test set. Each prompt contains a single fixed
few-shot example.

\textbf{Named-Entity Recognition.}
For CoNLL, the model generates a structured JSON object with fields for person,
organization, location, and miscellaneous entities. The fields are constrained
to task-specific vocabularies comprising 8{,}804 values. We report micro-F1 on
the 2{,}000-example test set.
Each prompt contains 10 fixed
few-shot examples, together with an input-specific subset of the entity
vocabulary precomputed before training.

For Gloss and CoNLL, the input-specific vocabulary subsets are precomputed
before training and evaluation and only inserted into the prompt at runtime;
no retrieval or vocabulary construction is performed during RL rollouts.

\vspace{-10pt}
\paragraph{Baselines.}
We compare GrammarRL with the frozen pretrained model under the same grammar constraints, using both grammar-constrained greedy and beam-search decoding. GrammarRL uses either the direct-only ($\lambda=0$), reverse-only ($\lambda=1$), or combined ($\lambda=0.5$) reward. Unless otherwise stated, GrammarRL is evaluated with greedy decoding; beam-search results are reported separately to distinguish gains from reinforcement learning from those obtained through inference-time search. Section~\ref{sec:inference-time} further compares GrammarRL with sampling-based inference-time methods.

\vspace{-10pt}
\paragraph{Training and Implementation Details.}
Unless otherwise stated, experiments use the default configuration in
Table~\ref{tab:hparams-task} in Appendix~\ref{app:implementation-details}.
All experiments use LoRA adaptation \citep{hu2021lora}, with pretrained weights
frozen as the reference model $\pi_{\mathrm{ref}}$. GrammarRL requires no
annotated targets: grammar-constrained candidates are generated from the
inputs and scored using the label-free reward defined in
Section~\ref{subsec:grammarl}.

Each rollout contains $N=3$ sampled trajectories and one
grammar-constrained beam candidate (beam width $B=3$), all trained jointly
with RLOO. The objective combines direct and reverse rewards
($\lambda=0.5$) with KL regularization toward $\pi_{\mathrm{ref}}$.
Further implementation details are provided in
Appendix~\ref{app:implementation-details}.

\subsection{Main Results}
\label{sec:exp-main-results}

We first evaluate GrammarRL across three model sizes against the frozen
pretrained model under the same grammar constraints. Table~\ref{tab:main-results-all-sizes}
reports the direct-only, reverse-only, and full GrammarRL objectives, with
grammar-constrained greedy decoding unless otherwise specified. Beam-search
results are reported separately to distinguish gains from reinforcement
learning from those obtained through inference-time search. Each configuration
is trained once and evaluated on 10 different (randomly sampled) test sets.

\begin{table}[h!]
\centering

\caption{ Results for Llama-3.2-1B-Instruct (1B), Llama-3.2-3B-Instruct (3B), and Llama-3.1-8B-Instruct (8B), with
$N{=}3$ and $B{=}3$. \textit{Baseline} denotes
the frozen pretrained model under the same grammar constraints, while
\textit{GrammarRL} denotes the model trained with our approach.
\textit{Reverse-only} and \textit{direct-only} use only the corresponding
reward, while \textit{GrammarRL} combines both with $\lambda{=}0.5$.
Rows marked \textit{beam} use beam search at inference time. Values are mean
$\pm$ standard deviation over 10 test sets, each
generated by randomly sampling the dataset with a different random seed.}
\label{tab:main-results-all-sizes}
\footnotesize
\begin{tabular}{llccc}
\toprule
Model & Config & Gloss (BLEU) & WoS (Hier.\ F1) & CoNLL (F1-micro) \\
\midrule
\multirow{6}{*}{1B}
& baseline (greedy)      & 0.342 $\pm$ 0.005 & 0.384 $\pm$ 0.006 & 0.435 $\pm$ 0.009 \\
& baseline (beam)        & 0.390 $\pm$ 0.018 & 0.385 $\pm$ 0.007 & 0.590 $\pm$ 0.006 \\
& GrammarRL (reverse only) & 0.554 $\pm$ 0.009 & 0.309 $\pm$ 0.006 & 0.548 $\pm$ 0.005 \\
& GrammarRL (direct only)  & 0.456 $\pm$ 0.005 & 0.288 $\pm$ 0.007 & 0.597 $\pm$ 0.010 \\
& GrammarRL              & \textbf{0.570 $\pm$ 0.014} & \textbf{0.394 $\pm$ 0.008} & 0.578 $\pm$ 0.006 \\
& GrammarRL (beam)       & 0.541 $\pm$ 0.013  & 0.391 $\pm$ 0.008 & \textbf{0.598 $\pm$ 0.006} \\
\midrule
\multirow{6}{*}{3B}
& baseline (greedy)      & 0.628 $\pm$ 0.007 & 0.442 $\pm$ 0.006 & 0.681 $\pm$ 0.006 \\
& baseline (beam)        & 0.663 $\pm$ 0.013 & 0.457 $\pm$ 0.006 & 0.819 $\pm$ 0.005 \\
& GrammarRL (reverse only) & 0.685 $\pm$ 0.009 & \textbf{0.526 $\pm$ 0.010} & 0.773 $\pm$ 0.005 \\
& GrammarRL (direct only)  & 0.648 $\pm$ 0.008 & 0.437 $\pm$ 0.007 & 0.787 $\pm$ 0.004 \\
& GrammarRL              & \textbf{0.686 $\pm$ 0.010} & 0.492 $\pm$ 0.008 & 0.812 $\pm$ 0.005 \\
& GrammarRL (beam)       & 0.671 $\pm$ 0.012  & 0.501 $\pm$ 0.009 & \textbf{0.835 $\pm$ 0.005} \\
\midrule
\multirow{6}{*}{8B}
& baseline (greedy)      & 0.611 $\pm$ 0.006 & 0.478 $\pm$ 0.007 & 0.752 $\pm$ 0.007 \\
& baseline (beam)        & 0.632 $\pm$ 0.006 & 0.512 $\pm$ 0.007  &0.857 $\pm$ 0.006 \\
& GrammarRL (reverse only) & 0.744 $\pm$ 0.006 & \textbf{0.622 $\pm$ 0.009} & 0.597 $\pm$ 0.008 \\
& GrammarRL (direct only)  & 0.681 $\pm$ 0.004 & 0.513 $\pm$ 0.007 & \textbf{0.861 $\pm$ 0.004} \\
& GrammarRL              & \textbf{0.754 $\pm$ 0.004} & 0.560 $\pm$ 0.006 & 0.787 $\pm$ 0.008 \\
& GrammarRL (beam)       & 0.745 ± 0.005 & 0.567 ± 0.006 & 0.801 $\pm$ 0.004 \\
\bottomrule
\end{tabular}
\end{table}

\vspace{-10pt}
\paragraph{GrammarRL versus the frozen baseline.}
The full GrammarRL objective outperforms greedy decoding of the frozen baseline on all three tasks and model
sizes, with an average gain of 9.8 points. The largest gains are on Gloss, where BLEU increases from $0.342$ to
$0.570$ for 1B, from $0.628$ to $0.686$ for 3B, and from $0.611$ to $0.754$ for 8B; on WoS and CoNLL the gains
range from 1.0 to 8.2 and from 3.5 to 14.3 points, respectively. Single-reward variants can be stronger in
individual settings (reverse-only on WoS for 3B and 8B, direct-only on CoNLL for 1B and 8B), but they can also
fall below the frozen baseline, as for reverse-only on CoNLL 8B ($0.597$ against $0.752$) and for both variants
on WoS 1B. The combined objective is the only configuration that improves over the baseline in every setting.
The reward ablation in Section~\ref{sec:ablations} further examines how the two signals contribute to these
gains.

\vspace{-10pt}
\paragraph{Greedy versus beam-search decoding.}
Beam search improves the frozen baseline across all
tasks, with the largest gains on CoNLL, where F1 increases from $0.435$ to
$0.590$ for 1B, from $0.681$ to $0.819$ for 3B, and from $0.752$ to $0.857$
for 8B. After GrammarRL, however, beam search does not consistently improve
performance: on CoNLL it provides only modest gains (e.g., from $0.578$ to
$0.598$ for 1B), while on Gloss it consistently reduces performance (e.g.,
from $0.570$ to $0.541$ for 1B). This may reflect the policy learning to
favor prefixes that lead to high-quality trajectories, reducing the need for
additional inference-time search.

\vspace{-10pt}
\paragraph{Training and generalization.}
Train and test performance remain close throughout training for the direct-only, mixed, and reverse-only
objectives, indicating that the reward-driven updates do not lead to severe overfitting
(Appendix~\ref{app:training-dynamics}, Figure~\ref{fig:curves}).

\vspace{-0.5em}

\subsection{Comparison with Inference-Time Methods}
\label{sec:inference-time}

Following \citet{loula2025syntactic}, we also compare GrammarRL (1B) with sampling-based methods that approximate
\eqref{eq:grammar-posterior} with $N{=}10$ particles. In their terminology, the grammar is an efficient
potential $\Phi_{\mathrm{eff}}$ folded into the proposal, while an expensive potential $\Phi_{\mathrm{exp}}$ is evaluated only on
the sampled particles; we use either the reverse likelihood $\Phi_{\mathrm{exp}}^{\mathrm{rev}}(y)=\pi_{\mathrm{ref}}(x\mid\tau(y))$
or task-specific potentials $\Phi_{\mathrm{exp}}^{\mathrm{task}}$ that check faithfulness to the source (e.g.\ entity
grounding, content-word coverage).
\emph{Sample-Rerank} keeps the sample with the highest $\Phi_{\mathrm{exp}}^{\mathrm{rev}}$; \emph{Full IS} also corrects the bias of
local masking with importance weights \citep{lipkin2025fast}; \emph{Full SMC} resamples the particles and reweights
every prefix with $\Phi_{\mathrm{exp}}^{\mathrm{task}}$. Rerank and IS thus use at inference time the signal that GrammarRL uses
only in training (Appendix~\ref{app:baselines}).

\begin{table}[t]
\centering
\footnotesize
\caption{Inference-time methods vs.\ GrammarRL (1B). Cost: generation relative to one greedy pass. Sampling
methods use $N{=}10$ particles and return the posterior mode. GrammarRL is reported with the default $\lambda{=}0.5$ and
with the best $\lambda$ per task in Table~\ref{tab:ablation-lambda} ($0.25$ for CoNLL and WoS, $0.75$ for Gloss); the
latter is selected on the same test seeds and is thus slightly optimistic. Mean $\pm$ std over 2 test sets of
2{,}000 examples; best value per column in bold.}
\label{tab:inference-time}
\setlength{\tabcolsep}{4pt}
\begin{tabular}{@{}llccc@{}}
\toprule
Method & Cost & Gloss (BLEU) & WoS (Hier.\ F1) & CoNLL (F1-micro) \\
\midrule
Constrained greedy            & $1\times$              & 0.339 $\pm$ 0.002 & 0.388 $\pm$ 0.000 & 0.434 $\pm$ 0.006 \\
Constrained beam ($B{=}3$)    & ${\approx}3\times$     & 0.380 $\pm$ 0.022 & 0.386 $\pm$ 0.002 & 0.586 $\pm$ 0.009 \\
Sample-Rerank ($\Phi_{\mathrm{exp}}^{\mathrm{rev}}$) & ${\approx}10\times$    & 0.346 $\pm$ 0.000 & 0.430 $\pm$ 0.009 & 0.488 $\pm$ 0.007 \\
Full IS ($\Phi_{\mathrm{exp}}^{\mathrm{rev}}$)       & ${\approx}10\times$    & 0.302 $\pm$ 0.002 & \textbf{0.442 $\pm$ 0.002} & 0.511 $\pm$ 0.010 \\
Full SMC ($\Phi_{\mathrm{exp}}^{\mathrm{task}}$)     & ${\approx}10\times$    & 0.236 $\pm$ 0.004 & n/a               & 0.578 $\pm$ 0.009 \\
\midrule
GrammarRL ($\lambda{=}0.5$)    & $1\times$ (+ training) & 0.554 $\pm$ 0.008 & 0.400 $\pm$ 0.010 & 0.577 $\pm$ 0.009 \\
GrammarRL (best $\lambda$)     & $1\times$ (+ training) & \textbf{0.559 $\pm$ 0.016} & 0.415 $\pm$ 0.004 & \textbf{0.621 $\pm$ 0.005} \\
\bottomrule
\end{tabular}
\vspace{-0.9em}
\end{table}

Table~\ref{tab:inference-time} shows that, at the cost of a single greedy pass, GrammarRL obtains the best score
on Gloss and CoNLL, ahead of methods that generate ten times more. On Gloss this holds with either $\lambda=0.5$ and $\lambda=0.75$
($0.554$ and $0.559$), while no sampling method improves meaningfully over greedy decoding and Full SMC falls well below
it ($0.236$ against $0.339$). On CoNLL, Full SMC with the grounding potential approaches beam search ($0.578$ vs. $0.586$, respectively);
GrammarRL surpasses both with $\lambda=0.25$ ($0.621$), whereas with the default $\lambda{=}0.5$ it is on par with
Full SMC ($0.577$) and below beam search, as in Table~\ref{tab:main-results-all-sizes}. WoS is the exception: Full IS and reranking with the reverse
likelihood outperform both GrammarRL variants ($0.442$ and $0.430$ against $0.415$ and $0.400$). On long WoS
abstracts the reverse likelihood is a sharp signal for selecting among candidates (Appendix~\ref{app:bl-fullis}),
and GrammarRL, which uses it only during training, recovers part of this benefit.

\vspace{-0.5em}

\subsection{Ablations}
\label{sec:ablations}

We next study the effect of the main design choices in GrammarRL: the number of sampled sequences used in each RLOO group, the beam width used during RL rollouts, and the mixing coefficient between the direct and reverse rewards. Unless otherwise stated, experiments use Llama-3.2-1B-Instruct, greedy decoding at evaluation time, and mean $\pm$ standard deviation over 10 sampled test sets of 2{,}000 examples.

\vspace{-10pt}
\paragraph{Rollout composition.}
GrammarRL is relatively insensitive to the number of sampled sequences: increasing $N$ from 3 to 9 changes
performance by at most $0.019$ on Gloss and $0.009$ on WoS and CoNLL. The training-time beam width has a larger
effect: $B{=}5$ slightly improves WoS and CoNLL, whereas $B{=}10$ reduces Gloss BLEU from $0.570$ to $0.470$,
suggesting that excessive search can shift the distribution of candidate trajectories in ways that hurt some
tasks. Full results are reported in Appendix~\ref{app:rollout-ablation}.

\vspace{-10pt}
\paragraph{Reward mixing coefficient.}
Table~\ref{tab:ablation-lambda} examines the relative contribution of the direct and reverse rewards, with $\lambda{=}0$ corresponding to direct-only and $\lambda{=}1$ to reverse-only training. On all three benchmarks, at least one intermediate mixture outperforms both single-direction objectives. The gains are largest on WoS, where performance increases from $0.288$ and $0.309$ for direct-only and reverse-only training, respectively, to $0.407$ at $\lambda{=}0.25$. This setting is also the best for CoNLL ($0.625$), while Gloss reaches its maximum at $\lambda{=}0.75$ ($0.576$); nevertheless, $\lambda{=}0.25$ still provides a substantial improvement over direct-only training on Gloss, from $0.456$ to $0.541$. Overall, the results show that the two reward directions are complementary, with intermediate reward settings outperforming single-direction training.

\begin{table}[t]
\centering
\footnotesize
\caption{ Ablation on the reward mixing coefficient $\lambda$, with $N{=}3$ and $B{=}3$ fixed. Experiments use Llama-3.2-1B-Instruct and greedy decoding. Values are mean $\pm$ std over 10 sampled test sets. $\lambda{=}0$ corresponds to direct-only training and $\lambda{=}1$ to reverse-only training. Intermediate mixtures improve over the single-direction objectives, although the best value depends on the task.}
\label{tab:ablation-lambda}
\begin{tabular}{lccc}
\toprule
Lambda value & Gloss (BLEU) & WoS (Hier.\ F1) & CoNLL (F1-micro) \\
\midrule
0 (direct only)   & 0.456 $\pm$ 0.005 & 0.288 $\pm$ 0.007 & 0.597 $\pm$ 0.010 \\
0.25              & 0.541 $\pm$ 0.004 & \textbf{0.407 $\pm$ 0.009} & \textbf{0.625 $\pm$ 0.008} \\
0.5               & 0.570 $\pm$ 0.014 & 0.394 $\pm$ 0.008 & 0.578 $\pm$ 0.006 \\
0.75              & \textbf{0.576 $\pm$ 0.013} & 0.367 $\pm$ 0.009 & 0.572 $\pm$ 0.005 \\
1 (reverse only)  & 0.554 $\pm$ 0.009 & 0.309 $\pm$ 0.006 & 0.548 $\pm$ 0.006 \\
\bottomrule
\end{tabular}
\end{table}

Neither reward dominates across all tasks, and the best mixture depends on the task. Appendix~\ref{app:Qualitative}
analyzes the error patterns induced by each reward.
Taken together, the ablations show that no single configuration is optimal for every task: moderate exploration
can help, whereas excessive training-time search ($B{=}10$ on Gloss) or an unbalanced reward mixture can hurt
performance. The best $\lambda$ depends on the task, but every task has an intermediate value that outperforms
both single-direction objectives, and the default $\lambda{=}0.5$ never falls below the frozen baseline
(Table~\ref{tab:main-results-all-sizes}).

\vspace{-0.3cm}

\section{Conclusion}
\label{sec:conclusion}
We introduced GrammarRL, a label-free reinforcement learning method that aligns language models to grammar
constraints. GrammarRL samples grammar-valid candidates and scores them with a bidirectional reward computed by a frozen reference model, which combines the likelihood of the output conditioned on the input and the likelihood of the input conditioned on the output. The resulting rewards are used to optimize the policy with RLOO. Across sign
language gloss translation, hierarchical classification, and named entity recognition, with Llama models from
1B to 8B parameters, GrammarRL improves constrained greedy decoding by 9.8 points on average (up to 22.8 BLEU) and, at
greedy-decoding cost, matches or exceeds beam search, and sampling-based methods that generate ten times more, on
two of the three tasks. The two
rewards are complementary: each one alone favors a different shortcut, such as uninformative outputs under the
direct reward and structural trap tokens under the reverse reward, whereas their combination improves over the
frozen baseline in every setting.

\vspace{-10pt}
\paragraph{Limitations and future work.}
The higher performance of our method comes at the cost of a (label-free) reinforcement learning phase, which might be computationally expensive. Moreover, changing the grammar would require re-running the training procedure. 
On CoNLL, beam
search on the frozen model remains slightly stronger than GrammarRL with greedy decoding. In this specific case, the gain in inference-time comes at the cost of a slight decrease in performance. %
A natural future direction is to combine GrammarRL with posterior-approximation methods rather
than treating them as alternatives: since these methods use grammar-constrained decoding as their proposal
distribution, the policy obtained after GrammarRL could serve as a better-informed proposal and reduce the
number of samples needed to approximate the target distribution. Other directions include adapting
$\lambda$ during training and extending GrammarRL to richer constraints such as programming languages.

\subsection*{AI use statement}
We used a large language model assistant during the preparation of this manuscript. It helped us correct
spelling and grammar, restructure the text and the appendices, check the consistency between text, tables,
and figures.
Generative AI tools were not used to design the method, run the experiments, or
produce the results. All AI-assisted text and code were reviewed by the authors and checked against our
experimental results. We take responsibility for the final content of this work, including text, claims,
or artifacts produced with the aid of generative AI.

\subsection*{Ethics statement}
All datasets used in this work are publicly available research benchmarks (ASLG-PC12, Web of Science, and
CoNLL-2003), and no new data involving human subjects were collected. The sign language task concerns
English-to-gloss translation on ASLG-PC12, whose glosses are generated automatically from English text
rather than produced by signers. Results on this benchmark should therefore not be taken as evidence of a
system ready to support communication with Deaf signers, and any such application would require the
involvement of the Deaf community. GrammarRL does not introduce risks beyond those of the underlying
pretrained language models, whose biases may be reflected in the constrained outputs.

\subsection*{Reproducibility statement}
Section~\ref{sec:methodology} describes the method. Appendix~\ref{app:calibration} reports the reward
scaling constants, Appendix~\ref{app:implementation-details} all training hyperparameters and LoRA settings,
and Appendix~\ref{app:prompts} the forward and reverse prompts. All datasets are publicly available, and all
backbones are publicly released Llama checkpoints. Results are averaged over 10 sampled test sets, each
drawn with a different random seed.

\bibliography{iclr2027_conference}
\bibliographystyle{iclr2027_conference}

\clearpage
\appendix

\section*{Appendix overview}
\addcontentsline{toc}{section}{Appendix overview}
Appendix~\ref{app:calibration} reports the reward scaling constants used in \eqref{eq:reward}.
Appendix~\ref{app:implementation-details} lists the training configuration and LoRA settings.
Appendix~\ref{app:prompts} describes the forward and reverse prompts.
Appendix~\ref{app:training-dynamics} shows training and test performance throughout training.
Appendix~\ref{app:rollout-ablation} reports the full ablation on the rollout composition.
Appendix~\ref{app:Qualitative} analyzes the error patterns induced by the direct and reverse rewards.
Appendix~\ref{app:baselines} describes the inference-time methods compared in Section~\ref{sec:inference-time}.

\FloatBarrier
\section{Reward Calibration}
\label{app:calibration}

The fixed constants $\sigma_{\mathrm{direct}}$ and $\sigma_{\mathrm{reverse}}$ in \eqref{eq:reward}
(Section~\ref{subsec:grammarl}) put the direct and reverse terms on comparable scales before they are blended
into a single reward. Rather than sharing one global pair of constants across all experiments, we measure
$\sigma_{\mathrm{direct}}$ and $\sigma_{\mathrm{reverse}}$ separately for each (task, model size, rollout
composition) combination, since both the model scale and the rollout composition (number of sampled
sequences $N$ and beam width $B$) change the natural spread of each term.

\begin{table}[!ht]
\centering
\footnotesize
\caption{Reward scaling constants $\sigma_{\mathrm{direct}}$ and $\sigma_{\mathrm{reverse}}$ per task, rollout
composition $(N,B)$ ($N$ sampled sequences, beam width $B$), and model size, measured on 100 held-out
development examples. Rows are grouped by the varied quantity: $N$ with $B{=}3$, then $B$ with $N{=}3$.
Shaded rows correspond to the default composition $(3,3)$ used in Table~\ref{tab:main-results-all-sizes}.
``--'' denotes a combination not used for that model size.}
\label{tab:calibration-all}
\setlength{\tabcolsep}{5pt}
\begin{tabular}{@{}llcccccc@{}}
\toprule
& & \multicolumn{2}{c}{1B} & \multicolumn{2}{c}{3B} & \multicolumn{2}{c}{8B} \\
\cmidrule(lr){3-4} \cmidrule(lr){5-6} \cmidrule(l){7-8}
Task & $(N,B)$ & $\sigma_{\mathrm{direct}}$ & $\sigma_{\mathrm{reverse}}$ & $\sigma_{\mathrm{direct}}$ & $\sigma_{\mathrm{reverse}}$
& $\sigma_{\mathrm{direct}}$ & $\sigma_{\mathrm{reverse}}$ \\
\midrule
\rowcolor{gray!12}
Gloss & $(3,3)$   & 1.1378 & 0.9731 & 0.6354 & 0.5493 & 0.4564 & 0.4442 \\
      & $(6,3)$   & 1.0693 & 1.0001 & --     & --     & --     & --     \\
      & $(9,3)$   & 1.1245 & 1.0229 & --     & --     & --     & --     \\
\addlinespace[2pt]
      & $(3,5)$   & 1.1147 & 0.9125 & --     & --     & --     & --     \\
      & $(3,10)$  & 1.1454 & 0.9453 & --     & --     & --     & --     \\
\midrule
\rowcolor{gray!12}
WoS   & $(3,3)$   & 0.5343 & 0.0265 & 0.2099 & 0.0166 & 0.1353 & 0.0140 \\
      & $(6,3)$   & 0.6387 & 0.0265 & --     & --     & --     & --     \\
      & $(9,3)$   & 0.6862 & 0.0295 & --     & --     & --     & --     \\
\addlinespace[2pt]
      & $(3,5)$   & 0.5346 & 0.0244 & --     & --     & --     & --     \\
      & $(3,10)$  & 0.5374 & 0.0269 & --     & --     & --     & --     \\
\midrule
\rowcolor{gray!12}
CoNLL & $(3,3)$   & 0.8047 & 0.2766 & 0.6405 & 0.2271 & 0.6253 & 0.1916 \\
      & $(6,3)$   & 0.8172 & 0.3190 & --     & --     & --     & --     \\
      & $(9,3)$   & 0.8652 & 0.3304 & --     & --     & --     & --     \\
\addlinespace[2pt]
      & $(3,5)$   & 0.8071 & 0.3053 & --     & --     & --     & --     \\
      & $(3,10)$  & 0.8378 & 0.2745 & --     & --     & --     & --     \\
\bottomrule
\end{tabular}
\end{table}

Each pair is the mean within-group standard deviation of the direct and reverse scores over rollout groups
sampled from \textbf{100 held-out examples per task} taken from the development set, disjoint from the
training, few-shot, and test data. The rollout groups use the same composition as the corresponding training run and are
generated by the frozen base model (no adapter). The constants are computed once and kept fixed throughout
training. Table~\ref{tab:calibration-all} reports the constants for every combination used in our experiments; ``--''
marks combinations that are not used for that model size.

We set $\lambda=0.5$, the midpoint of $\{0.25, 0.5, 0.75\}$, without tuning it per task, model size, or rollout
composition; Table~\ref{tab:ablation-lambda} reports the full sweep for the 1B model.

\FloatBarrier
\section{Implementation Details}
\label{app:implementation-details}

All experiments fine-tune LoRA adapters \citep{hu2021lora} with rank $r=64$, scaling factor $\alpha=128$, and
dropout $0.05$, applied to the attention and MLP projections. The pretrained weights are kept frozen and serve
as the reference model $\pi_{\mathrm{ref}}$. Table~\ref{tab:hparams-task} lists the default configuration used
in all experiments unless otherwise stated. The reward scaling constants for each task and backbone are
reported in Table~\ref{tab:calibration-all} (row $(3,3)$ for the default configuration).

\begin{table}[!ht]
\centering
\footnotesize
\caption{Default training configuration. Learning rate, gradient clipping threshold, and batch size do not
depend on the backbone. Values spanning all three columns are shared across tasks. Reward scaling constants
are reported in Table~\ref{tab:calibration-all}.}
\label{tab:hparams-task}
\setlength{\tabcolsep}{8pt}
\begin{tabular}{@{}lccc@{}}
\toprule
& \textbf{Gloss} & \textbf{WoS} & \textbf{CoNLL} \\
\midrule
\multicolumn{4}{@{}l}{\textit{Data}} \\
\quad Training examples           & 2k & 4k & 4k \\
\quad Test sets                   & $10\times2000$ & $10\times2000$ & $10\times2000$ \\
\quad Few-shot pairs / hints      & 20 / 50 & 1 / --- & 10 / 80 \\
\midrule
\multicolumn{4}{@{}l}{\textit{Optimization}} \\
\quad Prompts per step (batch)    & 4  & 8  & 8 \\
\quad Training steps (2 epochs)   & 1000 & 1000 & 1000 \\
\quad Learning rate               & $3.3\times10^{-7}$ & $7\times10^{-7}$ & $7\times10^{-7}$ \\
\quad Optimizer                   & \multicolumn{3}{c}{AdamW} \\ %
\midrule
\multicolumn{4}{@{}l}{\textit{Rollouts and reward}} \\
\quad Sampled sequences $N$       & \multicolumn{3}{c}{3} \\
\quad Beam width $B$ (training only) & \multicolumn{3}{c}{3} \\
\quad Sampling temperature        & \multicolumn{3}{c}{1.0} \\
\quad Max.\ new tokens            & 350 & 128 & 128 \\
\quad Reward mixing $\lambda$     & \multicolumn{3}{c}{0.5} \\
\quad KL coefficient $\beta$      & \multicolumn{3}{c}{0.02} \\
\midrule
\multicolumn{4}{@{}l}{\textit{LoRA}} \\
\quad Rank $r$ / scaling $\alpha$ & \multicolumn{3}{c}{64 / 128} \\
\quad Dropout                     & \multicolumn{3}{c}{0.05} \\
\quad Target modules              & \multicolumn{3}{c}{attention and MLP projections} \\
\midrule
\multicolumn{4}{@{}l}{\textit{Compute}} \\
\quad Hardware / time per run     & \multicolumn{3}{c}{1$\times$ H100 (80GB) / $\sim$5--18 hours\footnotemark} \\
\bottomrule
\end{tabular}
\end{table}
\footnotetext{Training time varies depending on the specific task, maximum generation length, and model size.}

\FloatBarrier
\section{Prompts}
\label{app:prompts}

We use few-shot chat prompts for all three tasks. Each prompt contains a system instruction, a fixed set of
demonstrations, and the input query. The demonstrations are selected once for each task and are reused
unchanged across queries. For Gloss and CoNLL, the prompts also contain query-specific \emph{hints}, which are
retrieved offline from an embedding index.

\paragraph{Forward (direct) prompt.}
The forward prompt maps an input query to the required structured output. It consists of (i) a task-specific
system instruction, (ii) a fixed set of input--output demonstrations (Table~\ref{tab:prompt-composition}), and (iii) the query as the final user turn.
For Gloss and CoNLL, the system instruction additionally contains hints retrieved for the current query using
\texttt{mxbai-embed-large-v1}: the nearest gloss terms for Gloss, and the nearest entity spans in the training
vocabulary for CoNLL. The hints are computed once offline and inserted into the prompt. Thus, the
demonstrations remain fixed across queries, whereas the hints change from one query to another.

\paragraph{Reverse prompt.}
The reverse prompt is constructed mechanically from the corresponding forward prompt. Each demonstration
$(u,a)$ is reversed to $(a,u)$, so that the model learns to reconstruct the original input from the structured
output. The forward system instruction is removed and replaced by a short task-specific inverse instruction,
and the retrieved hints are not included. Finally, the original query is replaced by a candidate output $y$,
from which the model must reconstruct the corresponding input.

Table~\ref{tab:prompt-composition} summarizes the composition of the forward prompts, and
Table~\ref{tab:prompt-system} decomposes each forward system instruction into its three parts and gives the
corresponding inverse instruction.

For WoS, the JSON keys \texttt{"parent"} and \texttt{"child"} hold the domain and area labels described in
Section~\ref{sec:exp-setup}.

\begin{table}[!ht]
\centering
\caption{Composition of the forward prompts. Few-shot demonstrations are selected once, deterministically,
without using text from the test sets, and are then reused for every query. Gloss and CoNLL additionally use
query-specific hints. The reverse prompt uses the same demonstrations with input and output swapped and
contains no retrieved hints.}
\label{tab:prompt-composition}
\footnotesize
\setlength{\tabcolsep}{5pt}
\begin{tabularx}{\textwidth}{@{}lccY@{}}
\toprule
Task & Few-shot pairs & Dynamic hints per query & Static system content \\
\midrule
Gloss    & 20 & 50 gloss terms              & --- \\
WoS      & 1  & ---                         & label hierarchy (7 parents, 134 children) \\
CoNLL    & 10 & $4\times20=80$ entity spans & --- \\
\bottomrule
\end{tabularx}
\end{table}

\begin{table}[!ht]
\centering
\footnotesize
\caption{Forward and reverse system instructions (abbreviated). Each forward instruction is composed of a fixed
\pInstr{task instruction}, a \pCtx{context block} (retrieved per query for Gloss and CoNLL, static for WoS), and
a \pOut{required output format}. Angle brackets mark slots whose content changes from one query to another. In
the reverse prompt, the forward system instruction and its context block are replaced by the inverse
instruction shown in the bottom half; the demonstrations are kept, with input and output swapped.}
\label{tab:prompt-system}
\footnotesize
\setlength{\tabcolsep}{5pt}
\renewcommand{\arraystretch}{1.15}

{\centering
\legendbox{cInstr}~\pInstr{instruction (fixed)}\qquad
\legendbox{cCtx}~\pCtx{retrieved / static context}\qquad
\legendbox{cOut}~\pOut{required output format}\par}
\medskip

\begin{tabular}{@{}
  >{\raggedright\arraybackslash}p{0.08\textwidth}
  >{\raggedright\arraybackslash}p{0.30\textwidth}
  >{\raggedright\arraybackslash}p{0.19\textwidth}
  >{\raggedright\arraybackslash}p{0.33\textwidth}@{}}
\toprule
\textbf{Task}
& \pInstr{\textbf{Instruction}}
& \pCtx{\textbf{Context}}
& \pOut{\textbf{Output format}} \\
\midrule
\multicolumn{4}{@{}l}{\textit{Forward system prompt}} \\
\midrule

Gloss
& \pInstr{Simplify the sentence into basic components. Separate components
  with a single space; preserve the meaning and structure of the original.}
& \slot{50 similar gloss terms} \newline
  {\scriptsize(retrieved)}
& \pOut{TERM TERM TERM \ldots} \newline
  {\scriptsize upper-case content words, space-separated} \\
\addlinespace[0.35em]

WoS
& \pInstr{Classify the abstract into one parent--child path of the
  hierarchy. Use labels from the hierarchy only.}
& \slot{label hierarchy} \newline
  {\scriptsize(7 parents, 134 children; static)}
& \pOut{\{"parent": ..., \newline "child": ...\}} \\
\addlinespace[0.35em]

CoNLL
& \pInstr{Extract named entities from the sentence. Copy the first span of
  each type verbatim; empty string if absent; emit the JSON object only.}
& \slot{4$\times$20 similar entity spans} \newline
  {\scriptsize(one line per type, retrieved)}
& \pOut{\{"person": ..., \newline "organization": ..., \newline
  "location": ..., \newline "misc": ...\}} \\

\midrule
\multicolumn{4}{@{}l}{\textit{Reverse system prompt} (replaces the forward instruction; no context block)} \\
\midrule

Gloss
& \multicolumn{3}{>{\raggedright\arraybackslash}p{\dimexpr0.82\textwidth+4\tabcolsep\relax}@{}}{\pInstr{Given the glossed form, reconstruct the
  original English sentence as it was actually written. Output only that
  sentence.}} \\
\addlinespace[0.2em]
WoS
& \multicolumn{3}{>{\raggedright\arraybackslash}p{\dimexpr0.82\textwidth+4\tabcolsep\relax}@{}}{\pInstr{Given the classification JSON,
  reconstruct the original abstract as it was actually written. Output only
  that text.}} \\
\addlinespace[0.2em]
CoNLL
& \multicolumn{3}{>{\raggedright\arraybackslash}p{\dimexpr0.82\textwidth+4\tabcolsep\relax}@{}}{\pInstr{Given the entity JSON, reconstruct the
  original sentence as it was actually written. Output only that sentence.}} \\

\bottomrule
\end{tabular}
\end{table}

\FloatBarrier
\section{Training Dynamics}
\label{app:training-dynamics}

\begin{figure}[!ht]
\centering
\includegraphics[width=\textwidth]{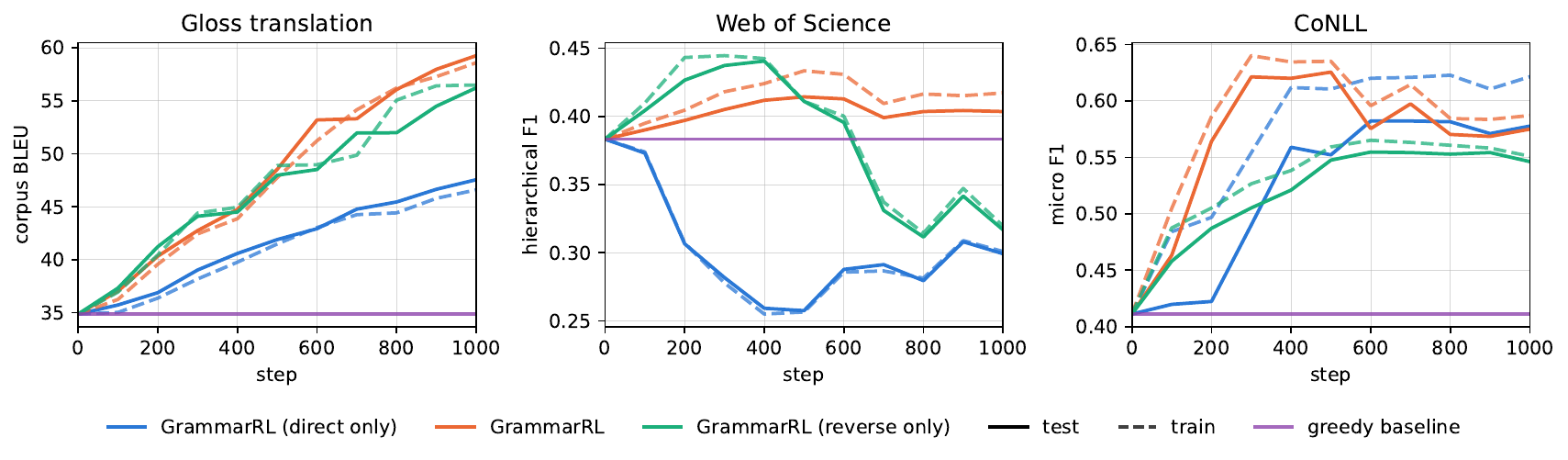}
\caption{Training and test performance throughout training for Llama-3.2-1B-Instruct with $N{=}3$ and $B{=}3$, comparing the GrammarRL direct-only ($\lambda{=}0$), GrammarRL ($\lambda{=}0.5$), and GrammarRL reverse-only ($\lambda{=}1$) objectives. Solid and dashed lines denote test and training performance, respectively. The relatively small train--test gaps throughout training indicate limited overfitting and suggest that the learned reward-driven behavior generalizes beyond the training examples.}
\label{fig:curves}
\end{figure}

Figure~\ref{fig:curves} shows training and test performance throughout training for the 1B model with $N{=}3$
and $B{=}3$, for the direct-only, mixed, and reverse-only objectives. The train and test curves
remain relatively close throughout training, with no pronounced divergence between the two splits. This
indicates that, under the considered setting, the reward-driven policy updates do not lead to severe
overfitting and that the learned behavior transfers reasonably well to unseen examples.

\FloatBarrier
\section{Rollout Composition Ablations}
\label{app:rollout-ablation}

Table~\ref{tab:ablation-rollout} reports the full ablation on the rollout composition summarized in
Section~\ref{sec:ablations}, with $\lambda{=}0.5$ fixed. The beam width $B$ affects only training-time rollouts
and is distinct from beam decoding at evaluation time.

\paragraph{Number of sampled sequences.}
With $B{=}3$ fixed, performance is relatively stable as $N$ increases from 3 to 9. Gloss shows the largest
variation, improving from $0.570$ to $0.589$, while WoS remains within a narrow range ($0.394$--$0.403$) and
CoNLL changes only slightly ($0.572$--$0.578$). GrammarRL is therefore relatively insensitive to the number of
rollout sequences in the explored range.

\paragraph{Beam width during RL rollouts.}
With $N{=}3$ fixed, increasing $B$ from 3 to 5 improves WoS from $0.394$ to $0.412$ and CoNLL from $0.578$ to
$0.587$, while Gloss decreases slightly from $0.570$ to $0.558$. Increasing the beam further to 10 causes a
marked regression on Gloss, which falls to $0.470$, whereas CoNLL remains relatively stable at $0.584$ and WoS at
$0.402$. A broader search space at training time is thus not always beneficial: excessive search may shift the
distribution of candidate trajectories in ways that hurt some tasks.

\begin{table}[!ht]
\centering
\caption{Ablation on the rollout composition, with $\lambda{=}0.5$ fixed (Llama-3.2-1B-Instruct, greedy
evaluation). Top: number of sampled sequences $N$ with $B{=}3$. Bottom: training-time beam width $B$ with
$N{=}3$. The shaded row is the default configuration $(N,B){=}(3,3)$, repeated in both blocks. Values are mean
$\pm$ std over 10 sampled test sets of 2{,}000 examples. Bold marks the best value in each block.}
\label{tab:ablation-rollout}
\footnotesize
\begin{tabular}{@{}lccc@{}}
\toprule
& Gloss (BLEU) & WoS (Hier.\ F1) & CoNLL (F1-micro) \\
\midrule
\multicolumn{4}{@{}l}{\textit{Sampled sequences $N$} ($B{=}3$)} \\
\rowcolor{gray!12}
\quad 3 & 0.570 $\pm$ 0.014 & 0.394 $\pm$ 0.008 & \textbf{0.578 $\pm$ 0.006} \\
\quad 6 & 0.584 $\pm$ 0.017 & \textbf{0.403 $\pm$ 0.008} & 0.577 $\pm$ 0.007 \\
\quad 9 & \textbf{0.589 $\pm$ 0.010} & 0.402 $\pm$ 0.009 & 0.572 $\pm$ 0.007 \\
\midrule
\multicolumn{4}{@{}l}{\textit{Beam width $B$} ($N{=}3$)} \\
\rowcolor{gray!12}
\quad 3  & \textbf{0.570 $\pm$ 0.014} & 0.394 $\pm$ 0.008 & 0.578 $\pm$ 0.006 \\
\quad 5  & 0.558 $\pm$ 0.015 & \textbf{0.412 $\pm$ 0.010} & \textbf{0.587 $\pm$ 0.006} \\
\quad 10 & 0.470 $\pm$ 0.014 & 0.402 $\pm$ 0.010 & 0.584 $\pm$ 0.006 \\
\bottomrule
\end{tabular}
\end{table}

\FloatBarrier
\section{Qualitative Analysis}
\label{app:Qualitative}

\begin{figure}[!ht]
\centering
\includegraphics[width=\linewidth]{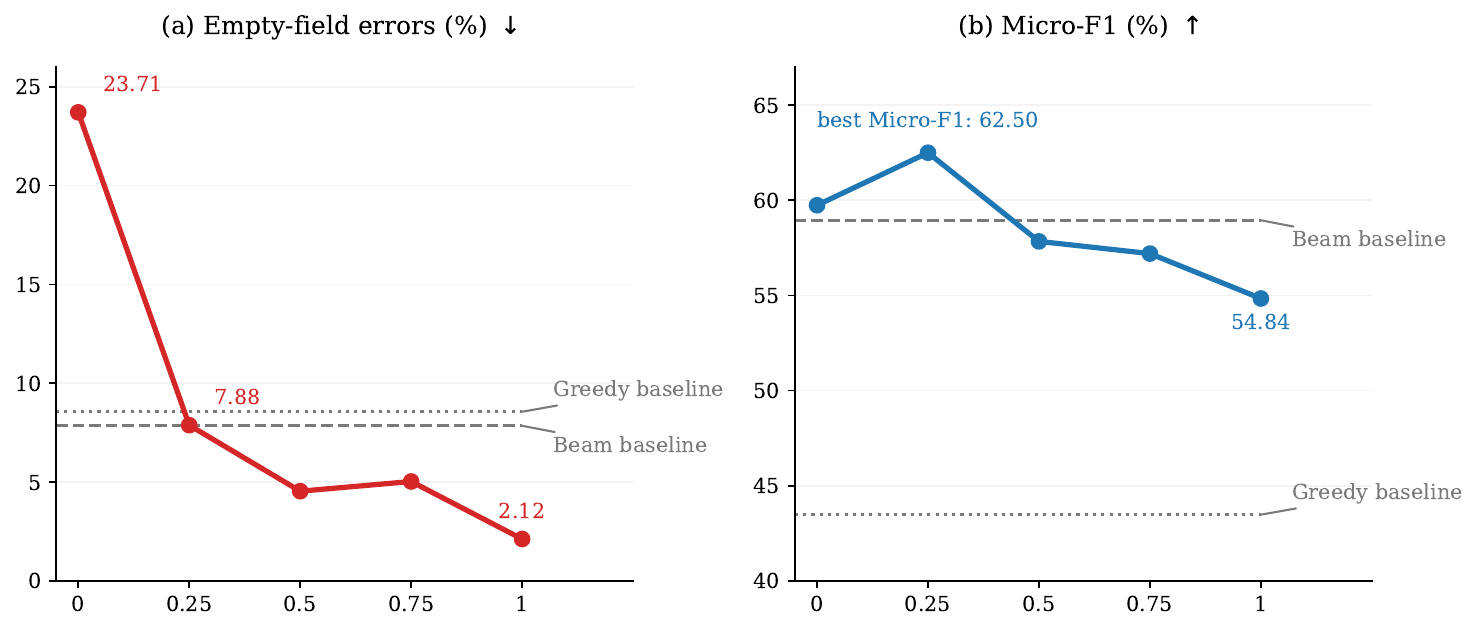}
\caption{Empty-field errors and micro-F1 on CoNLL as a function of the mixing coefficient $\lambda$
(Llama-3.2-1B-Instruct, $N{=}3$, $B{=}3$), averaged over 10 sampled test sets of 2{,}000 examples.
(a) Percentage of fields that are non-empty in the gold annotation but left empty in the prediction.
(b) Micro-F1. Horizontal lines mark the frozen model with greedy and beam-search decoding. Empty-field errors
decrease overall from 23.71\% at $\lambda{=}0$ to 2.12\% at $\lambda{=}1$, whereas micro-F1 peaks at
$\lambda{=}0.25$ (62.50).}
\label{fig:conll-lambda}
\end{figure}

We analyze how the mixing coefficient $\lambda$ changes the errors produced by GrammarRL, using
Llama-3.2-1B-Instruct with $N{=}3$ and $B{=}3$. As in the main experiments, all statistics are averaged over
10 sampled test sets of 2{,}000 examples each. Table~\ref{tab:reward-mechanism-gloss-lambda} counts spurious
tokens in Gloss predictions, and and Figure~\ref{fig:conll-lambda} measures missing entities in CoNLL
predictions together with micro-F1.

On Gloss, we track two kinds of spurious output. \texttt{X} is a gloss of the target vocabulary; we count its
occurrences as \emph{bare} \texttt{X}. Tokens with the \texttt{-DESC} suffix, such as \texttt{THEREFORE-DESC},
\texttt{THERE-DESC}, and \texttt{OFTEN-DESC}, are fragments obtained from the tokenization of longer glosses; we
refer to them as \emph{trap tokens}.

\begin{table}[!ht]
\centering
\caption{Trap tokens in Gloss predictions as a function of the mixing coefficient $\lambda$ (Llama-3.2-1B-Instruct,
$N{=}3$, $B{=}3$). Each entry is the mean number of occurrences over 10 sampled test sets of
2{,}000 examples. As $\lambda$ goes from 0 to 1, \texttt{THEREFORE-DESC} rises monotonically from 7 to 36 and
\texttt{THERE-DESC} from 33 to 140, whereas bare \texttt{X} falls monotonically from 115 to 51. \texttt{OFTEN-DESC} is rare for every $\lambda$ (at most one occurrence). Bold marks the
lowest value in each column.}
\label{tab:reward-mechanism-gloss-lambda}
\small
\begin{tabular}{@{}lcccc@{}}
\toprule
$\lambda$ & \texttt{THEREFORE-DESC} & \texttt{OFTEN-DESC} & \texttt{THERE-DESC} & bare \texttt{X} \\
\midrule
0 (direct only)   & \textbf{7}  & \textbf{0} & \textbf{33}  & 115 \\
0.25              & 13          & \textbf{0} & 46           & 84 \\
0.5               & 18          & 1          & 67           & 79 \\
0.75              & 24          & 1          & 90           & 54 \\
1 (reverse only)  & 36          & 1          & 140          & \textbf{51} \\
\bottomrule
\end{tabular}
\end{table}

\paragraph{Gloss.}
Changing $\lambda$ does not uniformly improve or degrade the output; it shifts the model between two kinds of
errors. Under the direct-only objective, bare \texttt{X} outputs dominate (115 occurrences). As the weight of
the reverse reward increases, bare \texttt{X} outputs decrease monotonically to 51, but structural trap tokens
become more frequent: \texttt{THERE-DESC} rises from 33 to 140 and \texttt{THEREFORE-DESC} from 7 to 36. The
reverse objective therefore does not simply remove degenerate behavior; it moves the model toward different
shortcuts. Intermediate values balance the two effects: at $\lambda{=}0.5$, bare \texttt{X} outputs fall to 79
while \texttt{THERE-DESC} (67) remains at less than half of its reverse-only count. This pattern is consistent
with Table~\ref{tab:ablation-lambda}, where the intermediate mixtures obtain the highest BLEU on Gloss
($0.576$ at $\lambda{=}0.75$ and $0.570$ at $\lambda{=}0.5$), above both direct-only ($0.456$) and reverse-only
($0.554$) training.

\begin{figure}[!ht]
\centering
\includegraphics[width=\linewidth]{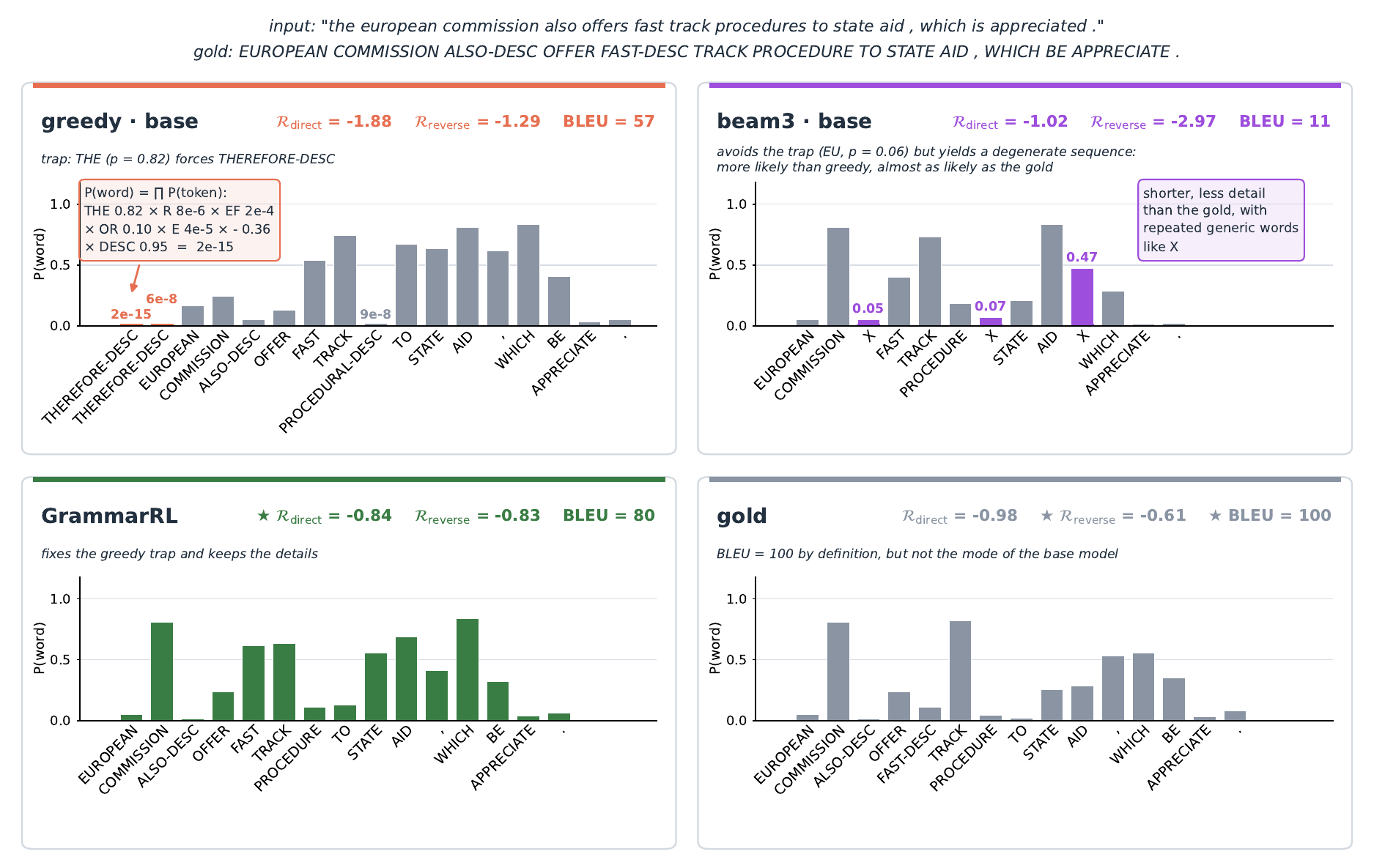}
\caption{Sequence likelihood versus correctness on one Gloss test example, chosen for readability
(outputs decoded deterministically). Each panel shows one output; panels marked \emph{base} decode with the
frozen model, and \emph{beam3} denotes beam search with $B{=}3$. Bars give the probability of each gloss
term under the frozen model before the grammar mask, computed as the product of its token probabilities.
The header reports the direct and reverse reward terms under the frozen model,
$\mathcal{R}_{\mathrm{direct}}$ and $\mathcal{R}_{\mathrm{reverse}}$ (Section~\ref{subsec:grammarl}), and
sentence-level BLEU; $\star$ marks the best value of each quantity.}
\label{fig:loglik-not-proxy}
\end{figure}

\paragraph{An example: likelihood versus correctness.}
Figure~\ref{fig:loglik-not-proxy} follows one Gloss test example through the three decoders. Greedy
decoding on the frozen model selects \texttt{THE}, the most likely first token ($p{=}0.82$). The grammar then
admits only glosses that begin with \texttt{THE}, and the model is forced to complete \texttt{THEREFORE-DESC}
through tokens it considers almost impossible. The trap occurs twice and accounts for most of the negative
log-likelihood of the sequence, yet the rest of the output is largely correct (BLEU 57). Beam search avoids
the trap by starting from \texttt{EU}, but it replaces three content terms with the generic gloss \texttt{X}
and drops others. The result is almost as likely under the frozen model as the gold
($\mathcal{R}_{\mathrm{direct}}$ of $-1.02$ against $-0.98$), although its BLEU is only 11: the direct reward
cannot distinguish this degenerate output from the correct one. The reverse reward can, because an output
with three \texttt{X} terms carries little information about the input, and beam search obtains by far the
lowest $\mathcal{R}_{\mathrm{reverse}}$ ($-2.97$). Across the four outputs, the reverse reward follows the
same order as BLEU, and so does the combined reward with the default $\lambda{=}0.5$. GrammarRL, trained
with this combined reward, avoids the trap and keeps the details of the sentence (BLEU 80).

\paragraph{CoNLL.}
Figure~\ref{fig:conll-lambda} shows a complementary effect on CoNLL. Under direct-only training, the model
leaves 23.71\% of the gold non-empty fields empty, almost three times as many as the frozen model with greedy
decoding (8.56\%). An empty field is always valid under the grammar but carries no information about the input,
so the direct reward alone increases the frequency of uninformative outputs. Adding even a small reverse weight
sharply reduces this behavior (7.88\% at $\lambda{=}0.25$), and reverse-only training leaves only 2.12\% of the
fields empty. Fewer empty fields do not imply higher F1, however: micro-F1 peaks at $\lambda{=}0.25$ (62.50) and
drops to 54.80 under reverse-only training, below direct-only training (59.70). This suggests that, without the
direct term, the model fills fields that should stay empty or fills them with incorrect spans.

\paragraph{Summary.}
Each reward used alone encourages a different shortcut that is valid under the grammar but detrimental to the
task: direct-only training produces outputs that carry little information about the input (empty fields on
CoNLL), whereas reverse-only training favors structural trap tokens on Gloss and, plausibly, incorrect spans on
CoNLL. Mixing the two rewards limits both failure modes, consistent with the higher scores of intermediate
values of $\lambda$ in Table~\ref{tab:ablation-lambda}.

\FloatBarrier
\section{Baselines and Inference-Time Methods}
\label{app:baselines}

Section~\ref{sec:inference-time} compares GrammarRL, evaluated with a single greedy grammar-constrained pass,
with methods that spend additional computation at inference time to approximate the grammar-conditioned
distribution $p_\theta(y\mid x,G)\propto p_\theta(y\mid x)\,\mathbb{I}[y\in\mathcal{L}(G)]$
\eqref{eq:grammar-posterior}, or to re-rank its samples. All methods use the same frozen
Llama-3.2-1B-Instruct, the same precomputed prompts and the same output language $\mathcal{L}(G)$; only the
decoding procedure, and the potentials that score partial or complete outputs, change.
Following \citet{loula2025syntactic}, the sampling-based methods target
$\pi_{\mathrm{ref}}(y\mid x)\,\Phi_{\mathrm{eff}}(y)\,\Phi_{\mathrm{exp}}(y)$ up to normalization, where the efficient
potential $\Phi_{\mathrm{eff}}(y)=\mathbb{I}[y\in\mathcal{L}(G)]$ is the grammar, enforced token by token, and the
expensive potential is either the reverse likelihood $\Phi_{\mathrm{exp}}^{\mathrm{rev}}$ (Appendix~\ref{app:bl-reverse})
or a task-specific potential $\Phi_{\mathrm{exp}}^{\mathrm{task}}$ (Appendix~\ref{app:bl-native}); with
$\Phi_{\mathrm{exp}}\equiv 1$ the target reduces to \eqref{eq:grammar-posterior}.
Table~\ref{tab:baselines-overview} summarizes the methods; the subsections below describe each of them, the
shared sampling machinery, and the efficient and expensive potentials that are used.

\subsection{Shared Machinery}
\label{app:bl-shared}

\paragraph{Output language.}
For the sampling-based methods we use \texttt{genlm-control} (v0.3.0), which constrains generation with a
byte-level automaton. We rebuild each task language as a deterministic finite automaton from the same files
used by the Grammar-LLM grammars: for CoNLL a JSON object with the four keys \emph{person, organization,
location, misc}, each value an element of the task vocabulary or the empty string; for WoS a JSON object
\texttt{\{"parent": $p$, "child": $c$\}} where the child ranges over the children of the parent just emitted
(145 taxonomy paths); for Gloss a space-separated sequence of the 16{,}120 gloss terms. Language equivalence is
checked on the gold outputs with an independent membership test. Generation ends with the model's end-of-turn
token, allowed only in accepting states, and is truncated at 128 new tokens as in the other methods.

\paragraph{Proposal and importance weights.}
Tokens are drawn by adaptive weighted rejection sampling (AWRS) \citep{lipkin2025fast} from the language model
restricted to the tokens the automaton admits. The result is a properly weighted particle: its weight corrects
the bias of local masking (a token that leads into a low-probability region of $\mathcal{L}(G)$ is not favored
just because the renormalization inflated its probability), so that the weighted particles target
$p_\theta(y\mid x,G)$ (times any expensive potential). The LM is sampled at temperature $1$ with the precomputed
few-shot prompt, encoded with the tokenizer's native chat template, exactly as in Grammar-LLM decoding.

\paragraph{Resampling.}
In SMC the particle population is resampled whenever the effective sample size $1/\sum_i \bar w_i^2$ falls below
$N/3$. With resampling disabled (Full IS) the $N{=}10$ particles are independent draws from the same proposal,
which is why Sample-Rerank can be computed from the very same samples as Full IS.

\paragraph{Output selection.}
A weighted population must be reduced to one output to be scored against the gold. We select the
\emph{posterior mode}: identical outputs are grouped, their normalized weights are summed, complete sequences are
preferred over sequences truncated at the token limit, and remaining ties are broken by the LM log-probability.
This is deterministic, and independent of whether the last SMC step happened to resample (after resampling all
weights are equal and a ``highest-weight particle'' would be arbitrary). We additionally log the
posterior-weighted metric used by \citet{loula2025syntactic}: the expected corpus metric when one output is drawn
per prompt from the particle posterior (20 draws).

\subsection{Constrained Greedy and Beam Search}
\label{app:bl-greedy}

These are the baselines of Table~\ref{tab:main-results-all-sizes}. The grammar is compiled into a pushdown
automaton whose admissible next tokens (with a token-boundary look-ahead engine that handles terminals spanning
several sub-word tokens) mask the logits; the masked distribution is renormalized \eqref{eq:constrained-policy}.
Greedy decoding takes the arg-max at every step. The beam variant runs beam search with three beams over the same
masked scores, so it can escape a locally attractive but globally poor prefix at the price of roughly three times
the compute. Both are deterministic and require no sampling framework.

\subsection{GrammarRL}
\label{app:bl-grammarrl}

GrammarRL is trained as described in Section~\ref{subsec:grammarl} and Appendix~\ref{app:implementation-details}
(RLOO over $N{=}3$ samples plus the top-1 beam hypothesis with $B{=}3$, KL coefficient $\beta{=}0.02$, 1000 steps,
reward constants from Table~\ref{tab:calibration-all}). At test time the model is decoded with the greedy
constrained decoder above, so its inference cost equals that of the frozen baseline. The comparison uses the
checkpoints at step 1000 and reports two settings: the default $\lambda{=}0.5$, used throughout the paper, and, for
each task, the best value of the mixing sweep in Table~\ref{tab:ablation-lambda} ($\lambda{=}0.25$ for CoNLL and
WoS, $\lambda{=}0.75$ for Gloss). The best value is selected with the same test protocol used for the comparison,
so the second row is slightly optimistic for GrammarRL; the gap between the two rows is large on CoNLL
($4.4$ points), small on WoS ($1.5$) and negligible on Gloss ($0.5$).

\subsection{Sample-Rerank}
\label{app:bl-rerank}

Sample-Rerank is the simplest use of an expensive potential and the control of the ladder of
\citet{loula2025syntactic} (grammar $\rightarrow$ Sample-Rerank $\rightarrow$ Full IS $\rightarrow$ Full SMC).
Draw $N{=}10$ independent samples from the grammar-constrained proposal, evaluate the potential $\Phi$ on each
complete sample, and return the sample maximizing it. Because $\Phi$ acts only after generation it cannot steer
it: if no sample is good, reranking cannot help. For $\Phi_{\mathrm{exp}}^{\mathrm{rev}}$ the score is the summed
log-likelihood, so $\arg\max_y \log\pi_{\mathrm{ref}}(x\mid\tau(y))$ is unchanged by the per-prompt constant that
separates this score from the length-normalized $\mathcal{R}_{\mathrm{reverse}}$ \eqref{eq:reverse}. Ties are
broken by the proposal weight. The reranked output is derived from the Full IS samples of the same prompt,
without generating a separate pool, which also makes the two methods paired.

\subsection{Full Importance Sampling (Full IS)}
\label{app:bl-fullis}

Full IS keeps the AWRS weights that correct local masking and multiplies them by the expensive potential
evaluated once on the finished sequence; no resampling takes place. The $N{=}10$ weighted particles approximate
$p_\theta(y\mid x,G)\,\Phi_{\mathrm{exp}}^{\mathrm{rev}}(y)$ and the output is the posterior mode. We call this variant
\emph{Full IS ($\Phi_{\mathrm{exp}}^{\mathrm{rev}}$)} because its potential is applied only at completion; the same potential inside an SMC
with resampling is ill-suited: a completion-only potential leaves unfinished sequences at weight one while
finished ones receive a large negative log-likelihood, so resampling removes finished particles and the
population drifts towards truncated outputs (we observed BLEU falling from 23 to 5 on Gloss in a pilot). With
strongly peaked weights (reverse log-likelihoods of a few hundred nats on long WoS abstracts) the effective
sample size is low, and Full IS then behaves close to Sample-Rerank. In Table~\ref{tab:inference-time}, importance weighting improves over
plain reranking on CoNLL and WoS but not on Gloss.

\subsection{Full SMC with Task-Specific Potentials}
\label{app:bl-smc}

Full SMC extends all particles one token at a time; after each step the expensive potential, acting as the critic,
reweights every particle by the ratio of its value after and before the step, and the population is resampled when
the ESS falls below $N/3$. This lets the potential penalize a particle as soon as the defect appears, instead of
waiting for the end of a long sequence. The expensive potential is the task-specific $\Phi_{\mathrm{exp}}^{\mathrm{task}}$ of
Appendix~\ref{app:bl-native}, evaluated on prefixes (completed fields or terms only) and on the complete output, with
soft (finite) penalties so that no particle is ever killed by a heuristic alone. Task-specific potentials exist for
CoNLL and Gloss; WoS has none, because its grammar already enforces the closed vocabularies and the hierarchy.

\subsection{Reverse (Noisy-Channel) Potential}
\label{app:bl-reverse}

The reverse potential is the likelihood, under the frozen model, of the input given the candidate output,
\begin{equation}
  \Phi_{\mathrm{exp}}^{\mathrm{rev}}(y)=\pi_{\mathrm{ref}}\!\left(x\mid\tau(y)\right)
   =\prod_{k=1}^{|x|}\pi_{\mathrm{ref}}\!\left(x_k\mid\tau(y),x_{<k}\right).
\end{equation}
The prompt $\tau(y)$ is the one used by the reverse reward \eqref{eq:reverse} (Appendix~\ref{app:prompts}), and
the source tokens are scored by teacher forcing in one forward pass with no grammar mask. This is the
noisy-channel potential \citep{yee2019simple}, with unit exponent and no free scale parameter. We deliberately do
\emph{not} reuse $\sigma_{\mathrm{reverse}}$ and $\lambda$: in GrammarRL they balance two signals inside an RLOO
advantage, whereas in a sampling target the natural unit of a log-potential is the log-likelihood itself. As a
consequence, the reverse log-likelihood, summed over $|x|$ tokens, is large in magnitude and the resulting
weights are sharp.

\subsection{Task-Specific Potentials}
\label{app:bl-native}

The task-specific potentials play the role of the domain-specific expensive potentials of \citet{loula2025syntactic} (e.g.\ plan
simulation, column validity): they use only the input $x$ (never the gold output), cost almost nothing (regular
expressions and string comparisons on the CPU), and encode one specific notion of \emph{faithfulness} to the
source. They are \emph{expensive} in the sense of \citet{loula2025syntactic} not because of the cost of a single
call, but because they are evaluated on the sampled particles and enter through the weights, rather than being
evaluated on every candidate next token and folded into the proposal. Each returns a log-weight $\log\Phi_{\mathrm{exp}}^{\mathrm{task}}(y)\le 0$, obtained as a sum of soft penalties; the
constants were fixed a priori and each rule was checked on 500 \emph{training} gold outputs before adoption,
never on the test sets. On a prefix only \emph{completed} units are judged (closed fields or terms followed by a
separator). Table~\ref{tab:native} lists the rules.

\begin{table}[t]
\centering
\scriptsize
\setlength{\tabcolsep}{3pt}
\caption{Task-specific potentials. $\log\Phi_{\mathrm{exp}}^{\mathrm{task}}$ is the sum of the penalties incurred; all penalties are soft,
so a violation lowers a particle's weight without ever removing it. On training gold, the CoNLL rules fire on none
of the 500 rows; for Gloss, $80.2\%$ of gold rows incur no penalty, and the mean gold penalty is $-0.17$ against
$-4.6$ for a shuffled control (source of row $i{+}1$ paired with the gloss of row $i$), about $27\times$ larger, so
the potential discriminates without being strict. WoS has no task-specific potential.}
\label{tab:native}
\begin{tabularx}{\linewidth}{@{}l l Z r Z@{}}
\toprule
\textbf{Task} & \textbf{Rule} & \textbf{Violation} & \textbf{Penalty} &
\textbf{Check on training gold (500 rows)} \\
\midrule
CoNLL & Grounding &
  An entity string is not a verbatim substring of the sentence. &
  $-3.0$ each & fires on 0\\[2pt]
CoNLL & Word boundary &
  The string occurs in the sentence but only inside longer words. &
  $-1.5$ each & fires on 0\\[2pt]
CoNLL & Duplicate &
  The same non-empty string is emitted in two different fields. &
  $-1.0$ per extra field & fires on 0\\
\midrule
Gloss & Precision &
  A completed gloss term is not grounded: after removing the \texttt{-DESC}/\texttt{-X} marker, its lemma matches
  no source word (suffix stripping and a dictionary lemmatizer) and it is not a closed-class function term
  (e.g.\ \emph{be, have, not, the}). &
  $-0.5$ each, at most 5 & fires on 14.6\% of rows \\[2pt]
Gloss & Coverage &
  A source content word (alphabetic, at least three letters, not a stop word) is not matched by any gloss term.
  Judged on the complete output only. &
  $-0.5$ each, at most 5 & gold covers $97.3\%$ of content words (shuffled control: $0.9\%$) \\[2pt]
Gloss & Degenerate run &
  A maximal run of at least three identical adjacent terms (a run of two, as in the gold \emph{BE BE}, is
  legitimate). &
  $-1.0$ per run & fires on 0\\
\bottomrule
\end{tabularx}
\end{table}

\paragraph{Relation to the reverse potential.}
Both express that the output must be faithful to the input: the task-specific rules test specific symptoms of
unfaithfulness (a hallucinated entity, an unsupported gloss term, an omitted content word) cheaply and locally,
whereas the reverse likelihood tests it generically and globally, at the cost of one forward pass of the frozen
model per candidate. The comparison therefore contrasts a hand-designed, task-specific critic with the same
label-free signal that GrammarRL uses for training.

\subsection{Evaluation and Scope}
\label{app:bl-scores}

All methods are scored with the same code as the main results: corpus BLEU for Gloss, hierarchical F1 for WoS,
and entity micro-F1 for CoNLL. For the sampling-based methods we report the score of the selected output
(posterior mode) and log the posterior-weighted metric and population diagnostics: the effective sample size, the
fraction of selected outputs that are complete, and the fraction of prompts on which all weights vanish (zero in
our pilots). The values in Table~\ref{tab:inference-time} are mean and standard deviation over two test sets of
2{,}000 prompts each, sampled as in Table~\ref{tab:main-results-all-sizes}; with two test sets the standard
deviation is only indicative. On the shared baselines the results agree with Table~\ref{tab:main-results-all-sizes}
(e.g.\ constrained greedy: $0.339$, $0.388$ and $0.434$ against $0.342$, $0.384$ and $0.435$).

The sampling methods use $N{=}10$ particles at temperature $1$ with a 1B-parameter model; they approximate the
same target and their behavior depends on the sharpness of the potential (Full IS and Sample-Rerank become nearly
deterministic for the reverse potential). The task-specific potentials are our own designs, kept deliberately generic (no
capitalization heuristics, no vocabulary derived from annotations) and validated on training gold only. Test
seeds are drawn from the part of the pool never used for training; on CoNLL, where that residual pool is small,
any two seeds share about a third of their examples, so the seed-to-seed spread is underestimated there.

\begin{landscape}
\begin{table}[p]
\centering
\scriptsize
\setlength{\tabcolsep}{3pt}
\caption{Methods compared with GrammarRL in Table~\ref{tab:inference-time}. All use Llama-3.2-1B-Instruct and the
same prompts. Efficient potentials $\Phi_{\mathrm{eff}}$ are cheap enough to be evaluated on every candidate next token and are
folded into the proposal; expensive potentials $\Phi_{\mathrm{exp}}$ are evaluated only on the sampled particles and
enter through the importance weights. $\Phi_{\mathrm{exp}}^{\mathrm{rev}}(y)=\pi_{\mathrm{ref}}(x\mid\tau(y))$
is the reverse likelihood of the input given the candidate (Appendix~\ref{app:bl-reverse}); the task-specific potentials
$\Phi_{\mathrm{exp}}^{\mathrm{task}}$ are defined in Appendix~\ref{app:bl-native}. Costs are relative to one greedy pass and refer to generation; reverse
scoring adds one forward pass of the frozen model per complete sample. GrammarRL uses the $N{=}3$, $B{=}3$ adapters,
with the default $\lambda{=}0.5$ and with the best $\lambda$ per task (Table~\ref{tab:ablation-lambda}).}
\label{tab:baselines-overview}
\begin{tabularx}{\linewidth}{@{}l Z Z Z Z Z Z@{}}
\toprule
\textbf{Method} & \textbf{What it does} & \textbf{Search at inference} &
\textbf{Efficient potential $\Phi_{\mathrm{eff}}$} & \textbf{Expensive potential $\Phi_{\mathrm{exp}}$} &
\textbf{Output selection} & \textbf{Cost} \\
\midrule
Constrained greedy &
  Mode of the locally renormalized masked policy $\pi^G$ \eqref{eq:constrained-policy}: one valid
  token at a time; no look-ahead. &
  None (single pass). &
  Grammar as a token mask (Grammar-LLM). &
  None. &
  Arg-max token at each step. &
  $1\times$ \\[2pt]
Constrained beam ($B{=}3$) &
  Searches for a higher-probability sequence under $\pi^G$; a mode-seeking
  approximation of $p_\theta(y\mid x,G)$. &
  Beam search, 3 hypotheses. &
  Same mask (Grammar-LLM). &
  None. &
  Top-scoring finished beam. &
  $\approx 3\times$ \\[2pt]
GrammarRL (ours) &
  Amortizes sequence-level preferences into the model: LoRA policy trained
  with RLOO on the label-free bidirectional reward, then decoded greedily. &
  None at test time (greedy). &
  Same mask, active in training and testing. &
  Used only in training: direct $+$ reverse reward from the frozen model
  \eqref{eq:reward}. &
  Arg-max token at each step. &
  $1\times$ (+ one-off training) \\[2pt]
\midrule
Sample-Rerank ($\Phi_{\mathrm{exp}}^{\mathrm{rev}}$) &
  Best-of-$N$ under an external score: $N$ i.i.d.\ grammar-constrained samples,
  keep the one with the highest reverse likelihood of the input. No weight
  correction, no resampling. &
  $N{=}10$ independent samples. &
  Grammar mask (proposal). &
  $\Phi_{\mathrm{exp}}^{\mathrm{rev}}$ (noisy-channel), at completion. &
  $\arg\max$ of the reverse score among complete samples. &
  $\approx N\times$ generation $+\,N$ scoring passes \\[2pt]
Full IS ($\Phi_{\mathrm{exp}}^{\mathrm{rev}}$) &
  Importance sampling from the grammar-constrained LM towards
  $p_\theta(y\mid x,G)\,\Phi_{\mathrm{exp}}^{\mathrm{rev}}(y)$; weights correct the myopia of local
  masking; no resampling. &
  $N{=}10$ independent weighted samples. &
  Grammar mask inside AWRS, with importance weight. &
  $\Phi_{\mathrm{exp}}^{\mathrm{rev}}$ (noisy-channel), applied once at completion. &
  Posterior mode: text with the largest total normalized weight. &
  Same as Sample-Rerank \\[2pt]
Full SMC ($\Phi_{\mathrm{exp}}^{\mathrm{task}}$) &
  Sequential Monte Carlo \citep{loula2025syntactic}: $N$ particles extended in parallel,
  resampled when the effective sample size drops, steered at every step by a
  task-specific expensive potential. &
  $N{=}10$ particles, resampling if $\mathrm{ESS}<N/3$. &
  Grammar mask inside AWRS, with importance weight. &
  $\Phi_{\mathrm{exp}}^{\mathrm{task}}$ (CoNLL, Gloss): grounding, coverage, boundary, repetition;
  on prefixes and at completion. &
  Posterior mode over particles. &
  $\approx N\times$ generation; potential evaluated on CPU \\
\bottomrule
\end{tabularx}
\end{table}
\end{landscape}

\end{document}